\documentclass[runningheads]{llncs}

\usepackage{eccv}

\usepackage{eccvabbrv}

\usepackage{graphicx}
\usepackage{booktabs}

\usepackage[accsupp]{axessibility}  % Improves PDF readability for those with disabilities.

\usepackage{hyperref}

\usepackage{orcidlink}

\usepackage{float}
\usepackage{todonotes} % for comments
 \usepackage{soul} % for highlights

\usepackage{xcolor,soul}
\colorlet{lightgray}{gray!20}
\sethlcolor{lightgray}
\usepackage{colortbl}
\usepackage[normalem]{ulem} % for tables
\usepackage{multirow}
\usepackage{multicol}% for tables
\useunder{\uline}{\ul}{} % for tables
\usepackage{diagbox}
\usepackage{slashbox}
\newcolumntype{C}[1]{>{\centering\arraybackslash\hspace{0pt}}p{#1}}
\usepackage{algorithm}
\usepackage{algpseudocode}
\algnewcommand{\Returns}{\textbf{Returns: }}
\usepackage{appendix}
\usepackage{tikz}
\usepackage{wrapfig}
\usepackage{graphicx}
\usepackage{xcolor}
\usepackage{varwidth}
\usepackage{environ}
\usepackage{xparse}
\usepackage[many]{tcolorbox}
\definecolor{bubblered}{RGB}{184,103,104}
\definecolor{bubblegreen}{RGB}{103,184,104}
\definecolor{bubblegray}{RGB}{241,240,240}
\newcommand*{\nolink}[1]{#1}
\begin{document}

\title{As Firm As Their Foundations} 
\subtitle{Can Open-Sourced Foundation Models be Used to Create Adversarial Examples for Downstream Tasks?}
% \subtitle{Crafting Adversarial Examples Across Multiple Downstream Tasks with Open-Sourced Foundation Models}
% \title{As Firm As its Foundations:
% Using an Open-Sourced Foundation Model to Generate Adversarial Examples for Typical Downstream Tasks} 

\titlerunning{As Firm As Their Foundations}

\author{Anjun Hu\inst{1} \and
Jindong Gu\inst{1} \and
Francesco Pinto\inst{1} \and\\
Konstantinos Kamnitsas\inst{1} 
\and
Philip Torr\inst{1} 
}

\authorrunning{A.~Hu et al.}

\institute{University of Oxford}

\maketitle

\begin{abstract}
% Foundation models are widely used as a basis for developing powerful machine learning systems. While they offer clear advantages for downstream task learning, they also introduces currently overlooked safety risks: 

Foundation models pre-trained on web-scale vision-language data, such as CLIP, are widely used as cornerstones of powerful machine learning systems. 
While pre-training offers clear advantages for downstream learning, it also endows downstream models with shared adversarial vulnerabilities that can be easily identified through the open-sourced foundation model.
In this work, we expose such vulnerabilities among CLIP's downstream models and show that foundation models can serve as a basis for attacking their downstream systems. In particular, we propose a simple yet alarmingly effective adversarial attack strategy termed Patch Representation Misalignment (PRM). Solely based on open-sourced CLIP vision encoders, this method can produce highly effective adversaries that simultaneously fool more than 20 downstream models spanning 4 common vision-language tasks (semantic segmentation, object detection, image captioning and visual question-answering). Our findings highlight the concerning safety risks introduced by the extensive usage of publicly available foundational models in the development of downstream systems, calling for extra caution in these scenarios.
  \keywords{AI safety, Adversarial Transferability, Foundation Models, Vision-Language Pretraining}
\end{abstract}

\section{Introduction}
\label{sec:intro}

Foundation models that combine both vision and language modalities \cite{jia2021ALIGN,li2019visualbert,lu2019vilbert,tan2019lxmert,chen2020uniter} are becoming increasingly popular, with CLIP \cite{Radford2021CLIP} standing out as a prime example. CLIP is extensively used by various downstream models, offering comprehensive cross-modality semantics to enhance a wide range of tasks such as open-vocabulary segmentation (OVS) \cite{bg-ovs-sparse-Luo2023SegCLIP,bg-ovs-dense-yu2023fcclip,bg-ovs-dense-xu2023san}, open-vocabulary object detection (OVD) \cite{bg-ovd-zhou2022detic,bg-ovd-wu2023cora,bg-ovd-zhao2022exploiting-VL-PLM}, image captioning (IC) \cite{bg-ic-mokady2021clipcap,bg-ic-li2023decap,bg-ic-nukrai2022CapDec} and visual question answering (VQA) \cite{bg-ic-vqa-shen2021clip-vil,bg-ic-vqa-liu2023llava,bg-ic-vqa-awadalla2023openflamingo}.

\begin{figure}[!t]
\includegraphics[width=\textwidth]{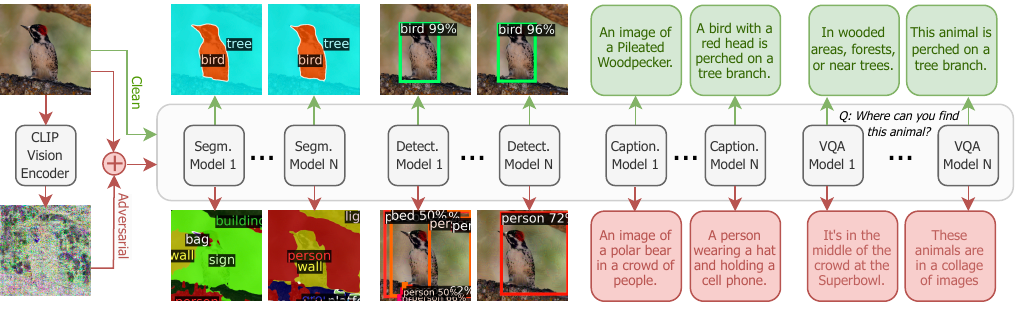}
\caption{Given a \textcolor{bubblegreen}{clean image} (top left), attackers can leverage the open-sourced CLIP vision encoder to find imperceptible \textcolor{bubblered}{input perturbations} (bottom left, magnified by 30$\times$ for visualisation) that distort CLIP's intermediate features. These perturbations are added to the original image to construct an \textcolor{bubblered}{adversarial sample} that can simultaneously fool many downstream models intended for various tasks: downstream models that are highly performant on clean samples (top row) suffer significant performance degradation (bottom row) under such attacks.\vspace{-10pt}}
\label{fig:teaser}
\end{figure}

While the usage of CLIP yields unprecedented performance improvements across many tasks, in this work, we show that such practice also introduces additional safety risks in downstream systems: since downstream models rely heavily on the semantics acquired during contrastive vision-language pretraining, adversarial perturbations that can disrupt such pre-trained semantic knowledge could significantly hamper these models' performance regardless of their architectural design or intended tasks. This implies that these models tend to share vulnerabilities to the same semantics-breaking adversarial examples, or that these adversaries demonstrate high \textit{transferability} across these downstream models. 
These adversaries pose significant safety risks as they make the downstream systems more prone to transfer-based attacks (i.e. more likely to be fooled by adversarial examples devised on models other than themselves).
Such risks are exacerbated by the fact that these highly transferable adversaries can be easily found through the publicly available CLIP model, rendering the downstream systems particularly susceptible to malicious exploitation.

To this end, we demonstrate the extent of such susceptibility by designing an adversarial attack strategy that creates highly transferable cross-task adversaries using only publicly available CLIP vision encoders. Notably, we show that an attacker with access to an off-the-shelf version of the CLIP vision encoder can substantially compromise the integrity of a wide range of downstream target models, including those that use an encoder with a different architecture than the one accessed by the attacker. Our contributions can be summarised as follows:

\begin{itemize}
    \item We show that adversarial examples built with off-the-shelf CLIP vision encoders can fool a wide range of downstream models in a task-agnostic manner, evidencing the propagation of adversarial vulnerability from open-sourced foundation models to downstream systems as a significant safety threat.
    \item We introduce Patch Representation Misalignment (PRM), an alarmingly effective adversarial attack strategy which seeks input perturbations that distort CLIP vision encoders' intermediate representations to undermine the predictive capability of downstream systems.
    \item Comprehensive experiments are conducted to show that this technique induces significant performance degradation in more than 20 target models spanning four common vision-language tasks, outperforming competing baselines \cite{bg-adv-madry2017pgd,bg-transfer-crosstask-naseer2018task,bg-transfer-crosstask-lu2020dispersion-reduction} by a significant margin.
\end{itemize}

While our experiments explore this phenomenon for CLIP, the extensiveness of our results and the alarming effectiveness of the proposed method calls for further efforts to understand whether this phenomenon extends to other foundational models. 
Our observations indicate that extreme caution should be taken to address the vulnerability introduced by the usage of foundational models in the development of novel systems.

\section{Related Works}

\subsubsection{Adversarial Transferability.}
Adversarial attacks introduce imperceptible perturbations to models' input data with the intention of causing degradation in model performance \cite{bg-adv-szegedy2013intriguing}. Early attack strategies such as \cite{bg-adv-goodfellow2014explaining,bg-adv-madry2017pgd} operate under the assumption of having white-box access to the target model. 
More recent works \cite{bg-adv-transfer-papernot2016transferability,bg-adv-transfer-liu2016delving} point out that such assumptions about target model accessibility are often unrealistic and subsequently show that it is possible to craft adversarial perturbations that \emph{transfer} across multiple models in a black-box fashion (i.e., without requiring access to the target's parameters or gradients). Such phenomenon of adversarial transferability, being the basis for black-box transfer attacks, has thus gained substantial research interest due to its significant implications for model safety \cite{bg-transfer-gu2023survey,bg-transfer-zhao2023revisiting}.
Crafting transfer-based attacks typically requires access to a surrogate model that shares some similarity with the target model, which could be in terms of model architecture, optimisation objectives, or training data. Having significant differences between the surrogate and the target model in these aspects may result in limited attack transferability \cite{bg-advseg-arnab2018robustness,methods-motivation-waseda2023closer} though many recent efforts have been made to foster stronger transferability despite these differences \cite{bg-transfer-surrogate-naseer2021improving,bg-transfer-generative-chen2023DiffAttack,bg-transfer-optim-zhang2023vit-token-gradient-regularizatione,bg-transfer-optim-qin2022boosting,bg-transfer-optim-wang2021enhancing,bg-transfer-optim-xiong2022stochastic,bg-transfer-optim-xu2022adversarially,bg-transfer-generative-naseer2019cross,bg-transfer-surrogate-zhao2023minimizing,bg-transfer-generative-naseer2021generating,bg-transfer-generative-yang2022C-GSP,bg-transfer-generative-song2018AC-GAN,bg-transfer-generative-poursaeed2018GAP}. 
In this work, we show that the extensive use of open-sourced foundation models in the development of other systems creates vulnerabilities that can be maliciously leveraged to improve attack transferability among downstream systems that exhibit significant differences in architectures and training paradigms.\\ 
\noindent Inspired by prior works on cross-task attacks \cite{bg-transfer-crosstask-naseer2018task,bg-transfer-crosstask-lu2020dispersion-reduction,bg-advdownstream-rezaei2019target} and feature-based attacks \cite{bg-transfer-feature-inkawhich2020perturbing,bg-transfer-feature-salzmann2021learning,bg-transfer-feature-ganeshan2019fda,bg-transfer-feature-huang2022defeat,bg-transfer-feature-huang2019enhancing,bg-transfer-feature-wang2021feature,bg-transfer-feature-salzmann2021learning}, we design a simple yet highly effective attack strategy that creates adversarial examples by finding input perturbations that induce maximal distortions in the surrogate model's internal representations. 
Different from prior works that focus on
distorting the magnitude or variance of model features,
we introduce a more refined objective that treats each patch independently and emphasises patch-wise directional misalignment between clean and adversarial features. 
This design serves to induce stronger semantic distortion in the CLIP feature space that densely covers all image regions and thereby produces more destructive adversaries compared to baseline methods.\vspace{-10pt}

\subsubsection{Foundation Models as Attack Surrogates.} Recently, there has been a growing focus on investigating and improving the robustness of vision-language foundation models \cite{bg-vlpadv-schlarmann2024robust,bg-vlpadv-zhou2023advclip,bg-vlpadv-noever2021reading,bg-foundationsafety-adv-carlini2023poisoning,bg-foundationsafety-qi2023fine,bg-vlpadv-zou2023universal,bg-foundationsafety-dong2023robust,bg-foundationsafety-tu2023many}. In particular, several studies have proposed strategies that use vision-language foundation models to devise attacks that transfer to other foundation models \cite{bg-transfer-vlp-zou2023universal,bg-transfer-vlp-shayegani2023jailbreak,bg-transfer-vlp-lu2023sga,bg-vlpadv-zhang2022towards,bg-transfer-vlp-zhao2024evaluating}. While existing works primarily concern the robustness and adversarial vulnerabilities of foundation models themselves, we focus on the fact that foundation models can serve as a basis for attacking their downstream systems. Our study extends the investigation on foundation model robustness to their downstream systems, which frequently engage in direct interactions with end users and can have a substantial negative impact on real-world applications if their integrity is compromised. 
% and highlights the the fact that foundation models can serve as a basis for attacking their downstream systems. 
\vspace{-22pt}

\subsubsection{Robustness of Pre-training and Downstream Models.} Several existing works have explored the inheritance of certain robustness properties in various transfer learning settings \cite{bg-advdownstream-yamada2022does,bg-advdownstream-hua2023initialization,bg-advdownstream-zhang2022remos,bg-advdownstream-wang2018great,bg-advdownstream-rezaei2019target}.
In particular, how various forms of weaknesses and limitations inherited from pre-training can influence downstream tasks \cite{bg-advdownstream-chen2024catastrophic,bg-advdownstream-nern2024transfer} have garnered growing research interest. Building upon this line of investigation, our study represents the first attempt to demonstrate the propagation of adversarial vulnerabilities from foundation models to downstream models, remarking the trade-off between utility and safety when using foundation models in the development of downstream systems.

\section{Methods}\label{sec:method}
In this section, we introduce our cross-task attack strategy termed Patch Representation Misalignment (PRM).
We first lay out the nomenclature for our problem setting and then motivate certain design choices including (1) attacking the intermediate features of a strong foundational vision model for cross-task generalisability and (2) using cosine similarity to diverge adversarial patch representations from their clean counterparts.
Following this, we delineate our approach in the latter half of this section.
 
\subsection{Preliminaries}\label{subsec:prelim}

\subsubsection{Notations and Threat Model.}
Given a clean input image sample $x$, a prediction ground truth $y$ and a target model $\mathcal{M}$ which is under attack, we consider an untargeted adversarial attack which aims at producing an adversarial sample $x' = x + \delta$ by adding an imperceptible perturbation $\delta$ to $x$ such that the target model $\mathcal{M}$ would produce an incorrect output prediction when given the adversarial sample $x'$ as input, i.e. $\mathcal{M}(x') \neq y$. To ensure that the attacks remain imperceptible, $\delta$ is often constrained by a perturbation budget $\epsilon$ such that the adversary stays within the $\epsilon$-neighbourhood of the original input. In this  work, we specifically consider $L_\infty$-norm bounded attacks (i.e. $\|x-x'\|_\infty\leq\epsilon$).
We assume that an attacker cannot query $\mathcal{M}$ nor access the parameters or gradients of $\mathcal{M}$. The attacker, however, has thorough knowledge about another model $\mathcal{F}$, referred to as a \textit{surrogate} model, which is expected to share some common characteristics as $\mathcal{M}$.
This allows the attacker to design algorithmic processes that can identify input perturbations that change the target network’s prediction.
Note that this threat model does not exclude the possibility of $\mathcal{F}$ and $\mathcal{M}$ having overlapping submodules (e.g. due to shared modules acquired from publicly available foundational models).\vspace{-10pt}

\subsubsection{Motivations for our Approach.}
Crafting adversarial samples that transfer between a wide range of multi-modal downstream systems across different tasks is non-trivial as these models exhibit substantial differences in terms of architecture, parameter values, and training paradigms.
While these differences often imply limited adversarial transferability \cite{bg-advseg-arnab2018robustness,methods-motivation-waseda2023closer}, we empirically show that the target models' mutual reliance on foundation models makes transfer attacks more feasible.
Drawing from potential challenges in designing a cross-task attack strategy, considering model characteristics, and gaining inspiration from prior works, we identify key aspects of an effective attack strategy as follows:
% (the efficacy of these design choices is empirically verified through ablation studies presented in Sec. \ref{subsec:ablations}):

% \noindent$\bullet$\textbf{What to attack}
\paragraph{Conducting feature-level attack for cross-task generalisability.} Output-level attacks that focus on the prediction (i.e. seek perturbations that mislead the network decisions by maximising training loss of a surrogate model) are dependent on the task-specific loss functions and therefore may demonstrate limited transferability across tasks and architecutres\cite{bg-transfer-crosstask-naseer2018task}. Hence, a feature-level attack strategy could be a more promising approach. Indeed, we may leverage the fact that downstream models frequently rely on CLIP vision encoders' intermediate features to devise feature-based attacks using CLIP vision encoders as surrogates. To show that well-designed feature-level attacks have more transferability potential than output-level attacks, we compare our method with task-specific output-level attack (training loss maximisation) baselines in Sec. \ref{subsec:main-results-discussions}.

% THIS NEEDS: a) if MSE $<$ Cos both on unaligned, then cite paper and use same argument, b) if MSE $>$ Cos on unaligned, then motivate Cos with clip alignemnt and MSE with realignment.
% Prior feature-attack strategies adopt global loss objectives which seek input perturbations that maximise global MSE between clean and adversarial features \cite{bg-transfer-crosstask-naseer2018task,bg-transfer-optim-zhang2023vit-token-gradient-regularizatione} or minimises global standard deviation of adversarial features \cite{bg-transfer-crosstask-lu2020dispersion-reduction}. While these approaches are effective to some extent, the gradients of these losses tend to be dominated by features with higher magnitudes or variances. As a result, they may overlook image regions with smaller magnitudes or variances (typically low spatial frequency regions), resulting in inadequate semantic distortion in these areas.
% To alleviate this issue, we take inspiration from CosPGD \cite{method-advseg-agnihotri2023cospgd}, an output-level attack for pixel-wise predictors that minimises cosine similarity between the distributions over the network predictions and ground truths, and hypothesize that an analogous approach can achieve a stronger feature-level attack. 

\paragraph{Taking a patch-wise approach to instil dense semantic distortions across all image regions.} When designing a cross-task attack strategy, we do not have prior knowledge about the vision granularity of the target task nor the potential regions of interest for a prediction. Therefore, an attack that can induce dense semantic distortions covering all image regions is more likely to successfully fool the target model.
We draw inspiration from existing output-level attacks for pixel-wise predictors \cite{method-advseg-agnihotri2023cospgd,method-advseg-gu2022segpgd} to design a feature-level attack that could densely affect all image regions. In particular, we adopt a patch-wise approach that independently diverts the representation of each adversarial image patch from the corresponding clean one by minimising the cosine similarity between the two, thereby densely covering the entire image with semantic distortions in a patch-by-patch manner.
To demonstrate the efficacy of this design, we compare the patch-wise approach with global alternatives equipped with different distance metrics ($L_2$ and angular distance) in Sec. \ref{subsec:main-results-discussions}-\ref{subsec:ablation-cosine-dim}.

% We hypothesize that, through the minimization of cosine similarity between clean and adversarial patch representations, this approach can independently ``flip'' the representation of each image patch, thereby introducing dense semantic distortion that spans the entire image. In other words, a cosine-based patch-wise objective could be less sensitive to spurious numerical traits that may otherwise prevent global losses from affecting certain image regions. \\
% \begin{figure}
%     \centering
%     \includegraphics[width=\textwidth]{images/Kurtosis.drawio.png}
%     \caption{While baseline methods are effective in texture-rich regions (R1 and R2), they tend to produce relatively benign, Gaussian-like perturbation patterns in regions that lack high-frequency features (R3) due to lower loss values in these regions. By employing magnitude-agnostic angular, our method can achieve stronger and semantically meaningful perturbations even in these difficult image regions.}
%     \label{fig:enter-label}
% \end{figure}
\paragraph{Emphasising directional information in the feature space.}
% The original CLIP model aligns language and vision using cosine similarities but such alignment was restricted to the global \texttt{CLS} token of the last layer. However, it has been demonstrated that fine-tuning or feature and architectural engineering can induce a similar form of language alignment for intermediate features \cite{bg-ovs-trainingfree-li2023clipsurgery,bg-ovs-trainingfree-sun2023clip,bg-ovs-sparse-mukhoti2023open}. 
CLIP's pre-training aligns image-level vision embeddings (\texttt{CLS} token of the last layer) and language using cosine similarity which emphasises directional information. It has been demonstrated that fine-tuning or feature and architectural engineering can induce a similar form of language alignment for CLIP's intermediate features \cite{bg-ovs-trainingfree-li2023clipsurgery,bg-ovs-trainingfree-sun2023clip,bg-ovs-sparse-mukhoti2023open}, suggesting that the orientation of these features may bear more semantic importance than their magnitude and variance. In other words, angular distance (between patch representations) could be a more appropriate metric than $L_p$ distance in the CLIP intermediate feature space.
Thus, driving directional dissimilarity between clean and adversarial features could potentially induce more potent distortions than existing feature attack approaches that focus on altering feature magnitude \cite{bg-transfer-crosstask-naseer2018task,bg-transfer-feature-salzmann2021learning} and variance \cite{bg-transfer-feature-salzmann2021learning}, providing another motivation to use cosine similarity minimisation as an adversarial objective intended for inducing sematic ``misalignment''.\\
To validate this design choice, we compare the efficacy of our method with $L_2$ distance-based and variance-based feature attack baselines in Sec. \ref{subsec:main-results-discussions}. 
In Sec. \ref{subsec:ablation-cosine-dim}, we also examine how the vision-language alignment pretraining objective may heighten the semantic importance of directional information in the features.

\subsection{Attack via Patch Representation Misalignment (PRM)}\label{subsec:loss}
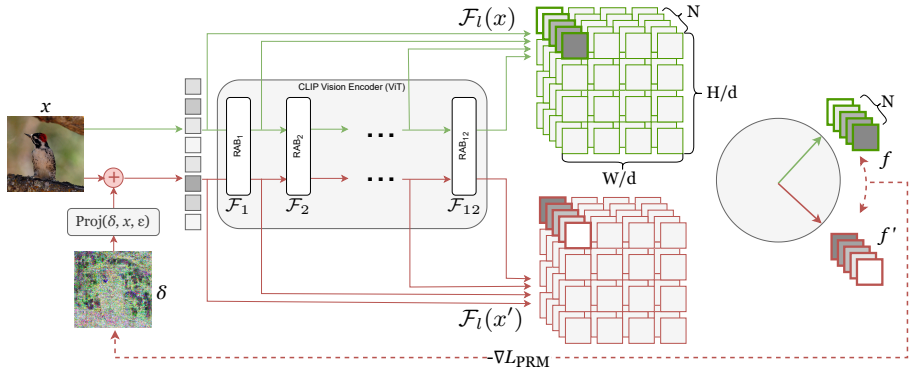
\begin{figure}[t]
    \centering
    \begin{tikzpicture}[]
    \node[anchor=north west,inner sep=0] at (-0.0, 0.0) {
    \includegraphics[width=.99\textwidth,trim={0 0cm 0 1cm},clip]{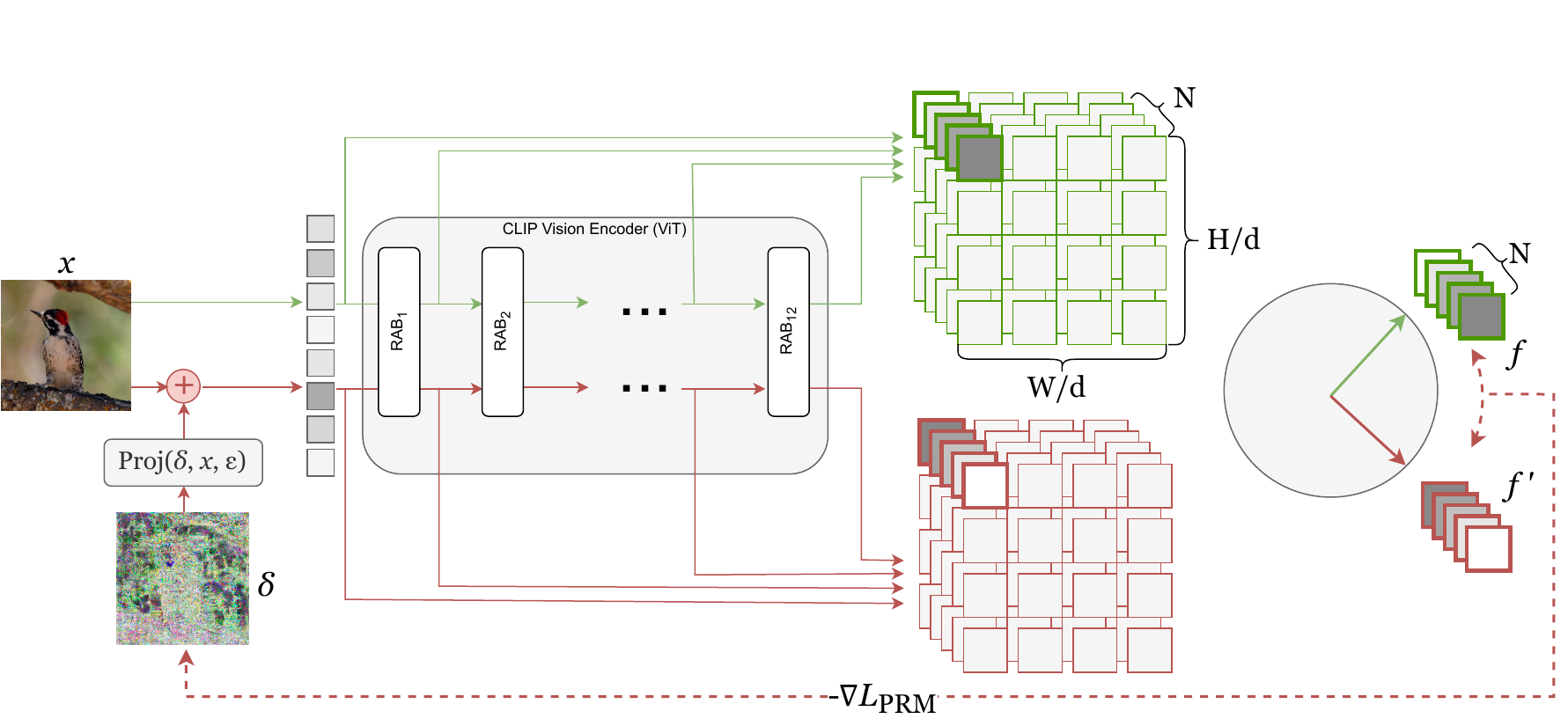}
    };
    \node[anchor=north west,inner sep=0] at (6, -0.3) {$\mathcal{F}_l(x)$
    };
    \node[anchor=north west,inner sep=0] at (6, -4.3) {$\mathcal{F}_l(x')$
    };
    \node[anchor=north west,inner sep=0] at (5.85, -2.85) {\scriptsize $\mathcal{F}_{12}$
    };
    \node[anchor=north west,inner sep=0] at (3.7, -2.85) {\scriptsize $\mathcal{F}_{2}$
    };
    \node[anchor=north west,inner sep=0] at (2.9, -2.85) {\scriptsize $\mathcal{F}_{1}$
    };
    \end{tikzpicture}
    \caption{Overview of our attack pipeline. A normal forward pass with clean input is marked in green whereas the forward pass of the adversarial sample is marked in red. Dashed line indicates the flow of loss gradients which are used to update the injected adversarial perturbation. 
    The loss objective minimises the cosine similarity between the adversarial representation of each patch (token) $f'$ and its clean counterpart $f$ along the embedding (ViT) or channel (CNN) dimension of the features. This approach individually diverts each patch representation (indicated by the reversed intensity of the top-left patch representation) to induce semantic distortions in all image regions.
    % This approach independently diverts the representation of each adversarial image patch $f'$ from the corresponding clean one $f$ by minimising the cosine similarity between the two to achieve strong semantic distortion in all image regions.
    }
    \label{fig:pipeline}
\end{figure}

As illustrated in Figure \ref{fig:pipeline}, our approach uses off-the-shelf CLIP vision encoders as surrogates and creates adversarial examples by iteratively optimising for input perturbations that induce maximal distortion of the surrogate model's internal representations.
Unlike prior work that focuses on disturbing feature magnitude or variance, our adversarial objective is designed to directionally misalign adversarial features from their clean counterparts in a patch-by-patch fashion to create maximal distortions of image semantics. 
% An alternative interpretation is to view it as causing a divergence in the distribution of each (normalized) representation of an adversarial patch from its clean counterpart.

Given a surrogate vision encoder $\mathcal{F}$, we denote intermediate features of a clean sample from encoder layer $l$ as $\mathcal{F}_l(x)$ and those of an adversarial sample as $\mathcal{F}_l(x')$. 
% Depending on the choice of CLIP vision encoder, each $\mathcal{F}_l$ can be a residual attention block \cite{dosovitskiy2020vit}, a ResNet \cite{he2016resnet} block or a ConvNeXt \cite{liu2022convnet} block.
For transformer-based vision encoders \cite{dosovitskiy2020vit} built with a series of residual attention blocks, $\mathcal{F}_l(x), \mathcal{F}_l(x') \in \mathbb{R}^{1+\lceil\frac{HW}{d^2}\rceil \times N}$. Each of them can be expanded as a set of $N$-dimensional embeddings for individual tokens $\mathcal{F}_l(x) = \{f^0_l, f^1_l, \dots, f^{\lceil\frac{HW}{d^2}\rceil}_l\} = \{f^p_l \in \mathbb{R}^{N} | p\in[0,\lceil\frac{HW}{d^2}\rceil]\}$, corresponding to representation of image patches of size $d$ and one global \texttt{CLS} token. 
We treat each token (patch or \texttt{CLS}) as an independent sample and minimise the cosine similarity between the adversarial embedding $f'$ of each token and its clean counterpart $f$, which drives adversarial token representations away from their clean counterparts and thereby misaligns them with the correct semantics:
\begin{align}\label{eq:loss-prm}
    \mathcal{L}_\mathrm{PRM} = \sum_{l \in L} \sum_{p=0}^{\lceil\frac{HW}{d^2}\rceil } \frac{{f}^p_l \cdot {f'}^p_l}{\|{f}^p_l \| \|{f'}^p_l\|}
\end{align}
% As discussed in \ref{subsec:prelim}, misaligning intermediate patch embeddings could be an effective way to create semantic distortions. While we cannot directly contrast intermediate patch embeddings with the text embeddings since they are not fully aligned in the first place, .

It is straightforward to adapt this loss for convolutional vision encoders built upon ResNet \cite{he2016resnet} or ConvNeXt \cite{liu2022convnet} blocks. In this case, clean and adversarial intermediate features from layer $l$ are $\mathcal{F}_l(x), \mathcal{F}_l(x') \in \mathbb{R}^{\lceil\frac{H}{d}\rceil \times\lceil\frac{W}{d}\rceil \times N} $ where $N$ is the number of filters (channels) and $d$ is a downsampling factor.
Rather than viewing the latent features as $N$ \textit{feature maps}, we regard them as a collection of $N$-dimensional \textit{descriptors} for each spatial element in the features. We reuse the notation of $f \in \mathbb{R}^N$ to denote these descriptors due to the conceptual similarity between these descriptors and the spatial token embeddings in ViTs: both serve as representations for image patches (overlapping patches in CNNs and disjoint patches in the ViTs).
A similar minimisation of cosine similarity between clean and adversarial descriptors can be done via Eq.\eqref{eq:loss-prm} where the $f$s are indexed by their spatial locations. 

We include pseudocode in Appendix \nolink{\ref{apx:algorithm}} and remark that our attack does not require access to task-specific surrogate models or ground truth annotations. It can be applied alongside other transferability-enhancing techniques based on input diversity \cite{bg-transfer-aug-DIM,bg-transfer-aug-SIM,bg-transfer-aug-TIM,bg-transfer-aug-RDI}, optimisation \cite{bg-transfer-optim-qin2022boosting,bg-transfer-optim-wang2021enhancing,bg-transfer-optim-xiong2022stochastic,bg-transfer-optim-xu2022adversarially,bg-transfer-optim-zhang2023vit-token-gradient-regularizatione} and surrogate refinement \cite{bg-transfer-surrogate-huang2023tsea,bg-transfer-surrogate-Chen_2023_ICCV,bg-transfer-surrogate-zhao2023minimizing,bg-transfer-surrogate-naseer2021improving,bg-transfer-surrogate-malik2022adversarial} to further amplify its effectiveness (e.g. the effect of input scale diversity is examined in Sec. \ref{subsec:ablation-scale} and Appendix \nolink{\ref{apx:subsec:scale}}). 
% For instance, we choose to incorporate random input rescaling to address Obstacle 7 identified in Sec. \ref{subsec:prelim}.

\section{Experiments and Evaluation}\label{sec:experiments}
% In this section, we present the results of our transfer attacks aimed at a comprehensive range of CLIP downstream models trained for various tasks.

\subsection{Experimental Setup}\label{subsec:setup}
\subsubsection{Baselines.} Since there are no well-established and directly comparable baselines that conduct task-generalisable attacks with non-CNN surrogates, we choose three adversarial loss objectives originally developed in other contexts as our baseline methods. We apply minor adaptations to these baselines to fit our problem setting and test each of them on two surrogate choices.
\begin{enumerate}
    \item \textbf{Task-Specific Baseline: Training Loss Maximisation} is the conventional approach taken by the original PGD\cite{bg-adv-madry2017pgd} method which requires a task-specific surrogate and access to ground truths for computing the surrogate loss. We choose two open-vocabulary segmentation models SAN \cite{bg-ovs-dense-xu2023san} and DeOP \cite{bg-ovs-dense-Han2023DeOP} as the task-specific surrogate model and use the corresponding negated surrogate training loss (denoted as $-\mathcal{L}_\mathrm{SAN}$ or $-\mathcal{L}_\mathrm{DeOP}$ in tables) as the adversarial loss objective. 
    \item \textbf{Cross-Task Baseline 1: Neural Representation Distortion (NRD)} \cite{bg-transfer-crosstask-naseer2018task} is a cross-task attack strategy based on maximising perceptual dissimilarity between clean and adversarial samples, measured by $L_2$ distance between VGG-16 conv3.3 features \cite{johnson2016perceptual}.
    To make it a comparable baseline applicable to CLIP vision encoders, we simply adapt their objective $\mathcal{L}_\mathrm{NRD} = -\sum_{l \in L}  \| \mathcal{F}_l(x) - \mathcal{F}_l(x') \|_2$ to ViT and ConvNeXt intermediate features and attack features from all encoder layers (setting $L$ to be all layers).
    \item \textbf{Cross-Task Baseline 2: Dispersion Reduction (DR)} \cite{bg-transfer-crosstask-lu2020dispersion-reduction} is a cross-task attack strategy that manipulates adversarial features from one selected CNN layer to become ``less dispersed'' by minimising their standard deviation. 
    % Such manipulations are observed to have lasting effects on distorting activations in subsequent model layers deeper than the one being attacked, thereby destroying the predictive capability of victim models. 
    To make it a comparable baseline, we similarly adapt their loss objective 
    $\mathcal{L}_\mathrm{DR} = - \sum_{l \in L} \sigma(\mathcal{F}_l(x'))$ to ViT and ConvNeXt intermediate features and attack features from all layers. 
\end{enumerate}
\vspace{-10pt}
\subsubsection{Implementation Details.}
We set a perturbation budget of $\epsilon = 8/255$ across all experiments and consider untargeted attacks. All experiments are boosted with random resizing as input augmentation \cite{bg-transfer-aug-RDI,bg-transfer-aug-DIM} with scaling factors ranging in (0,75, 1.25) and optimized for $N=250$ (converged) iterations with AdamW\cite{loshchilov2017adamw} optimizer at a learning rate of $\alpha=0.5/255$. 
We present our strategy using either ViT-B/16 or ConvNeXt-L CLIP vision encoders as surrogates with the same loss objective in Eq.\ref{eq:loss-prm}.
Attackable layers $L$ are set to all 12 Residual Attention Blocks for the former or all ConvNeXtBlocks in the 3 ConvNeXt stages (excluding stem) for the latter.
Using two surrogates helps us to (1) validate the efficacy of our loss design on different architectures; (2) examine transferability in scenarios where the vision encoder architecture of the surrogate matches or mismatches that of the target models' vision encoder.

\begin{figure}[t]
    \centering
    \includegraphics[width=.49\textwidth,trim={3.7cm 1.9cm 5.1cm 1.9cm},clip]{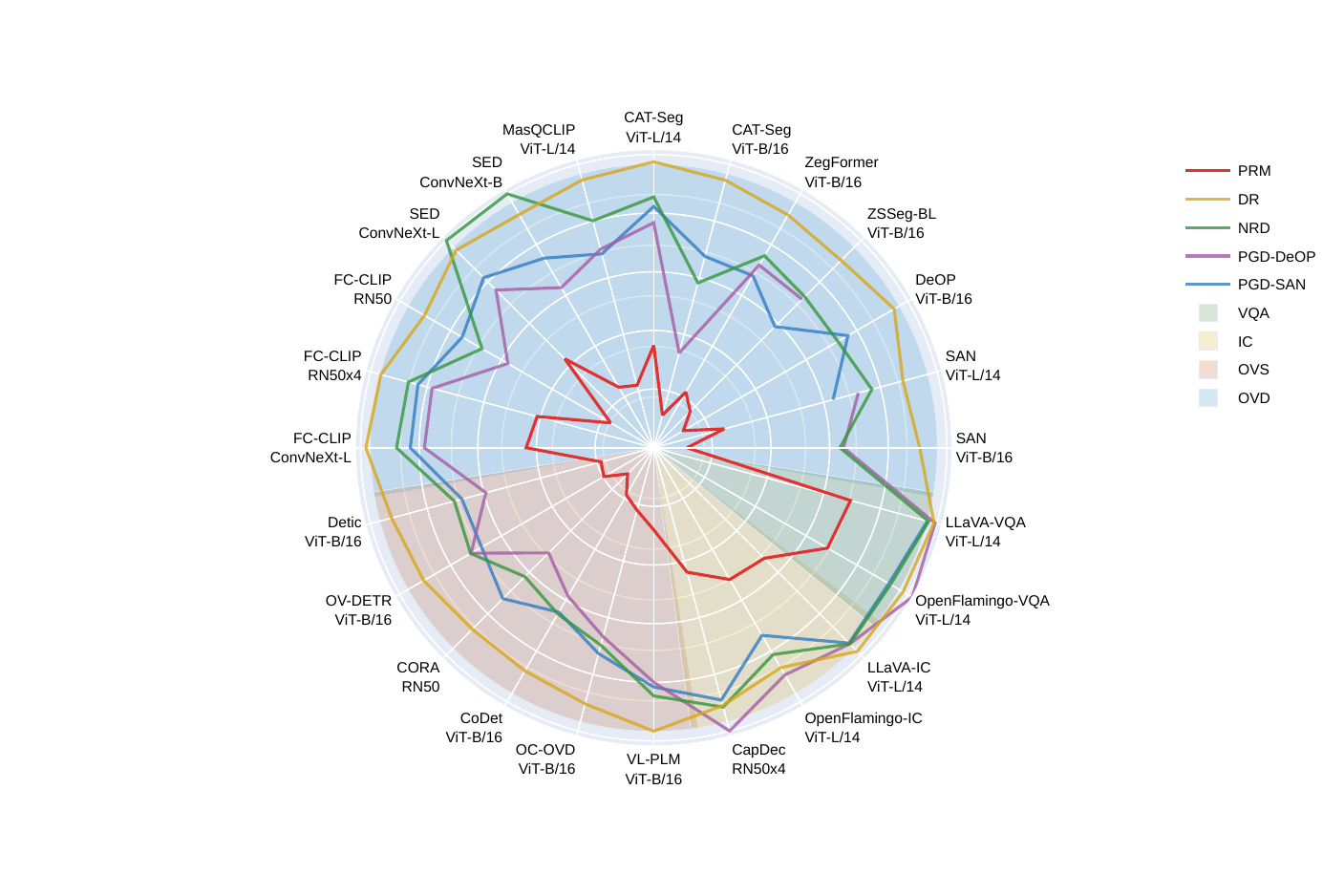}
    \includegraphics[width=.49\textwidth,trim={3.7cm 1.9cm 5.4cm 1.9cm},clip]{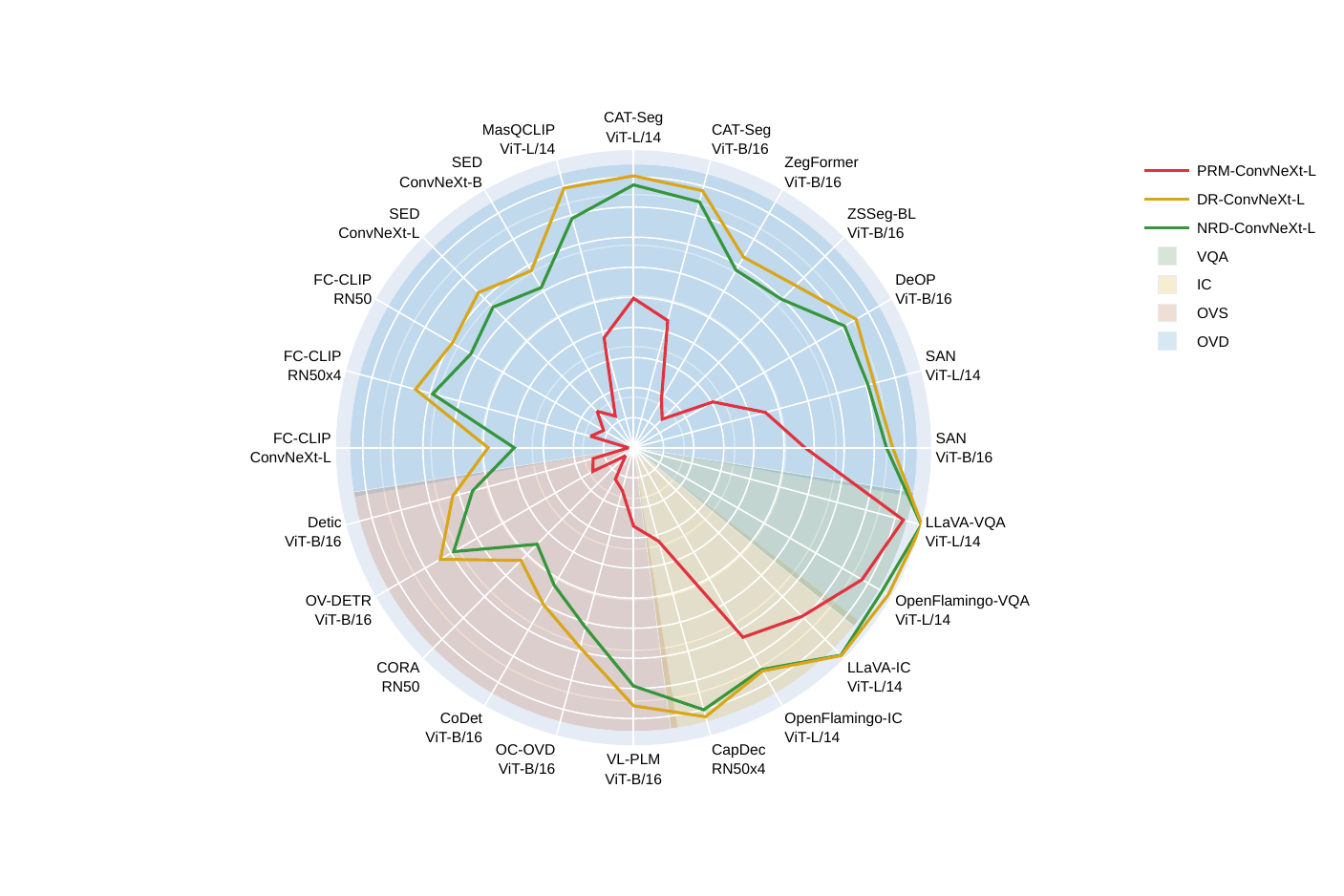}
    \includegraphics[width=.9\textwidth]{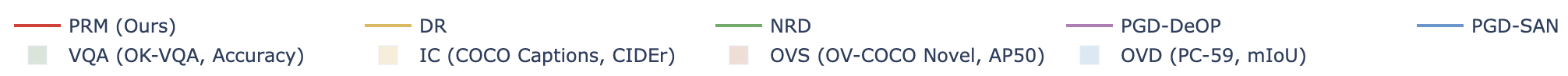}
    \caption{Normalised target model performance (model performance metrics under adversarial attacks divided by metrics on clean samples) of various attack strategies. \\
    \textbf{Left:} using ViT-B/16 (or task-specific baselines that use ViT-B/16 backbone) as surrogates. \textbf{Right: }using ConvNeXt-L as surrogates. PRM (red line) outperforms baseline methods by a significant margin across all tasks with both surrogate choices.\\
    The radius of the outer circles represents model performance on clean samples (unitary normalised metrics). Each attack strategy corresponds to a line. Each task is indicated by a differently coloured sector (datasets and metrics used for each task are detailed in the $2^\mathrm{nd}$ line of the legend).  Target model names are annotated on the periphery of the circles. 
    White-box scenarios in surrogate loss maximisation baselines are excluded. \vspace{-10pt}
    } 
    \label{fig:radial-plots}
\end{figure}

\subsection{Tasks, Datasets and Metrics}

\noindent\textbf{Target Task 1: Open Vocabulary Semantic Segmentation (OVS).}
% We specifically consider semantic segmentation in our experiments, but note that the efficacy of the attacks is likely generalizable to other segmentation tasks such as panoptic and instance segmentation. 
The efficacy of our attack on OVS target models is evaluated using two popular semantic segmentation benchmarks: Pascal Context \cite{dataset-mottaghi2014pascal-context} and COCO-Stuff \cite{dataset-caesar2018coco-stuff}.
COCO-Stuff is a large-scale dataset for semantic segmentation consisting of 118k training images, 5k validation images and 41k testing images with 171 annotated classes. Pascal Context is another semantic segmentation benchmark containing 5k training images and 5k validation images; we use the 59-class label set (PC-59) in our evaluation.
% It is widely used in open-vocabulary semantic segmentation under the cross-dataset evaluation setting where a model trained on COCO-Stuff is directly tested on the PC validation set.
% OVS target models considered in our experiments are trained on the COCO-Stuff training set. Hence, while both datasets measure a model's segmentation capability, the performance on PC-59 additionally represents a model's open-vocabulary generalizability. 
We attack eleven SoTA OVS models of which eight are densely supervised by pixel-level annotation (SAN\cite{bg-ovs-dense-xu2023san}, DeOP\cite{bg-ovs-dense-Han2023DeOP}, ZSSeg-Baseline\cite{bg-ovs-dense-xu2022zssegbaseline}, ZegFormer\cite{bg-ovs-dense-ding2021dZegFormer}, CAT-seg\cite{bg-ovs-dense-cho2023catseg}, MasQCLIP\cite{bg-ovs-dense-xu2023masqclip}, SED\cite{bg-ovs-dense-xie2023sed}, FC-Clip\cite{bg-ovs-dense-yu2023fcclip}) and three are weakly supervised by image-level labels or text (SegCLIP\cite{bg-ovs-sparse-Luo2023SegCLIP}, TCL\cite{bg-ovs-sparse-cha2022tcl}, SimSeg\cite{bg-ovs-sparse-yi2023simseg}). Target model performance on the two validation sets is measured with the mean of class-wise intersection over union (mIoU) metric. Results are reported in Table \ref{tab:ovs-pc59} for PC-59 and \nolink{Table \ref{tab:ovs-coco} (Appendix \ref{apx:quantitative})} for COCO-Stuff.

% COCO-Stuff is a large-scale dataset for semantic segmentation consisting of 118K training images, 5K validation images and 41K testing images with 171 annotated classes. OVS target models considered in our experiments are all trained on the COCO-Stuff training set. Pascal Context (PC) is a dataset for semantic segmentation which contains 5K training images and 5K validation images. We use the 59-class label set in our evaluation. It is widely used in open-vocabulary semantic segmentation under the cross-dataset evaluation setting, where a model trained on COCO-Stuff is directly tested on the PC validation set. While both datasets serve to measure a model's segmentation capability, the performance on PC additionally represents a model's open-vocabulary generalizability. 

\vspace{5pt}
\noindent\textbf{Target Task 2: Open Vocabulary Object Detection (OVD).}
For evaluating attack efficacy on OVD target models, we use the OV-COCO \cite{dataset-lin2014microsoftcoco} dataset which consists of 107761 training images and 4836 validation images. We adopt the widely used base-novel split of class labels, where 48 base classes are used for training and an additional 17 novel classes are introduced during inference.
We attack six OVD target models (VL-PLM\cite{bg-ovd-zhao2022exploiting-VL-PLM}, Detic\cite{bg-ovd-zhou2022detic}, CORA\cite{bg-ovd-wu2023cora}, CoDet\cite{bg-ovd-ma2023codet}, object-centric-ovd\cite{bg-ovd-Hanoona2022Bridging-object-centric-ovd,bg-ovd-Maaz2022Multimodal-object-centric-ovd}, OV-DETR\cite{zang2022OV-DETR}). Target model performance is measured on the OV-COCO validation set using the AP50 (the average precision at an IoU of 50\%) metric. We report model performance on base and novel classes separately in Table \ref{tab:ovd}. 

\vspace{5pt}
\noindent\textbf{Target Task 3: Image Captioning (IC).}
We generate adversarially perturbed versions of COCO Captions \cite{chen2015coco-caption} dataset and evaluate attack efficacy on three victim models (CapDec\cite{bg-ic-nukrai2022CapDec}, OpenFlamingo\cite{bg-ic-vqa-awadalla2023openflamingo}, LLaVA\cite{bg-ic-vqa-liu2023llava}) using the 5000-sample test set in Karpathy splits.
As shown in Table \ref{tab:ic-vqa}, we evaluate caption quality with several commonly adopted metrics: BLEU score (B@1, B@4) \cite{cap-metric-papineni2002bleu}, METEOR (M) \cite{cap-metric-denkowski2014meteor}, ROUGE-L (R-L) \cite{cap-metric-lin2004rouge-l} and CIDEr (C) \cite{cap-metric-vedantam2015cider}.

\vspace{5pt}
\noindent\textbf{Target Task 4: Visual Question Answering (VQA).}
We generate adversarially perturbed versions of OK-VQA \cite{dataset-ren2015coco-qa} and evaluate attack efficacy on two victim models (OpenFlamingo\cite{bg-ic-vqa-awadalla2023openflamingo}, LLaVA\cite{bg-ic-vqa-liu2023llava}), as shown in Table \ref{tab:ic-vqa}.
Target model performance is measured with 5046 validation questions from OK-VQA using the VQA accuracy metric.

\newcommand{\bluebf}[1]{\textbf{\color{blue} #1}}

\begin{table}[t] 
\fontsize{6}{8}\selectfont
\centering
\caption{Transfer attack efficacy on OVS target models. Model performance is measured in mean Intersection over Union (\textbf{mIoU}) where a lower metric indicates higher attack efficacy. Each row (S) represents a surrogate-loss configuration while each column (T) corresponds to a target model. Vision encoder backbones used by each model are specified below model names. We divide surrogate configurations into two groups (separated by \underline{\underline{double lines}}) mirroring the left and right radial plots in Figure \ref{fig:radial-plots}: the first group uses ViT encoders (or task-specific model with ViT backbones) as surrogates; the second group uses ConvNeXt encoders as surrogates.
\hl{Light grey cells} indicate scenarios where the source and target models have matching vision encoder architectures. Excluding  {\color{gray} white-box attacks} where source and target models are identical, the \bluebf{best} attack strategy for each target model within each group is highlighted in \bluebf{blue}.} \label{tab:ovs-pc59}
\begin{subtable}{\textwidth}
% \caption{Pascal Context 59.}
\begin{tabular}
{cc|C{.95cm}|C{.95cm}|C{1cm}|C{1.40cm}|C{1.40cm}|C{.95cm}|C{.95cm}|C{1.4cm}}
\specialrule{.8pt}{1pt}{1pt}
 &  & \multicolumn{2}{c|}{SAN} & DeOP & ZSSeg-BL & ZegFormer & \multicolumn{2}{c|}{CAT-Seg}  & MasQCLIP\\
 \cline{3-10}
 & \multirow{-2}{*}{$\mathcal{L}_\mathrm{attack}$} & ViT-B & ViT-L & ViT-B & ViT-B & ViT-B& ViT-B & ViT-L  & ViT-L \\
 \cline{2-10}
\multirow{-3}{*}{\backslashbox[8mm]{S}{T}} 
&  Clean & 54.07 &  57.70 &  48.65 &  46.92 &  41.27 &  57.4 &  61.97 &  58.65\\
 \hline
 \hline
SAN & $-\mathcal{L}_\mathrm{SAN}$ &\cellcolor[HTML]{EFEFEF}{\color{gray} 7.12}&   36.54 &\cellcolor[HTML]{EFEFEF}37.21 &\cellcolor[HTML]{EFEFEF}27.42 &\cellcolor[HTML]{EFEFEF}27.94  &\cellcolor[HTML]{EFEFEF}38.84 & 51.00& 40.15\\
DeOP & $-\mathcal{L}_\mathrm{DeOP}$ &\cellcolor[HTML]{EFEFEF}35.00 &   41.68 &\cellcolor[HTML]{EFEFEF}{\color{gray} 4.57} &\cellcolor[HTML]{EFEFEF}33.59 &\cellcolor[HTML]{EFEFEF}29.69 &\cellcolor[HTML]{EFEFEF}19.25 & 47.57& 41.10 \\
 \hline
 & DR &\cellcolor[HTML]{EFEFEF}49.06 &   50.78 &\cellcolor[HTML]{EFEFEF}46.07 &\cellcolor[HTML]{EFEFEF}42.42 &\cellcolor[HTML]{EFEFEF}37.83 &\cellcolor[HTML]{EFEFEF}54.30 & 60.44& 55.44 \\
 & NRD &\cellcolor[HTML]{EFEFEF}34.34 & 44.45 &\cellcolor[HTML]{EFEFEF}35.29 &\cellcolor[HTML]{EFEFEF}34.17 &\cellcolor[HTML]{EFEFEF}31.20&\cellcolor[HTML]{EFEFEF}33.46 & 53.01  & 47.00\\
\multirow{-3}{*}{\begin{tabular}[c]{@{}c@{}}CLIP\\ ViT-B\end{tabular}} & \textbf{PRM} &\cellcolor[HTML]{EFEFEF}\bluebf{6.13} &\bluebf{14.37} &\cellcolor[HTML]{EFEFEF}\bluebf{ 5.69} &\cellcolor[HTML]{EFEFEF}\bluebf{8.26} &\cellcolor[HTML]{EFEFEF}\bluebf{9.05} &\cellcolor[HTML]{EFEFEF}\bluebf{6.64} & \bluebf{21.63} & \bluebf{12.93} \\
 \hline
 \hline
 & DR & 46.57 & 47.76 & 41.54 & 35.77 & 30.18  & 50.84 & 56.00& 52.39\\
 & NRD & 45.45 & 46.54 & 39.39 & 32.71 & 28.12 & 48.62 & 54.23& 46.28 \\
\multirow{-3}{*}{\begin{tabular}[c]{@{}c@{}}CLIP\\ {CNeXt-L}\end{tabular}} & \textbf{PRM} & \bluebf{30.81} & \bluebf{26.18} & \bluebf{14.83} & \bluebf{6.31} & \bluebf{7.64} & \bluebf{25.15} & \bluebf{30.79}& \bluebf{22.20} \\
\specialrule{.8pt}{1pt}{1pt}
\end{tabular}
\begin{tabular}{cc|c|c|C{.95cm}|c|c|c|C{.95cm}|C{.9cm}|C{.9cm}}
\specialrule{.8pt}{1pt}{1pt}
 &   & \multicolumn{2}{c|}{SED} & \multicolumn{3}{c|}{FC-CLIP} & SegCLIP & TCL & \multicolumn{2}{c}{SimSeg}\\
 \cline{3-11}
 & \multirow{-2}{*}{$\mathcal{L}_\mathrm{attack}$}& CNeXt-B & {CNeXt-L} & RN50 & RN50$\times$64 & {CNeXt-L}& {ViT-B} & {ViT-B} & {ViT-S} &{ViT-B} \\
 \cline{2-11}
\multirow{-3}{*}{\backslashbox[8mm]{S}{T}} 
& Clean &  57.67 &  60.87 &  51.80 &  55.15 &  56.99  &  24.25 &  33.9  &  25.81  &  26.18 \\
 \hline
  \hline
SAN & $-\mathcal{L}_\mathrm{SAN}$ & 42.95 & 49.94 & 39.07 & 45.92 & 47.32 &\cellcolor[HTML]{EFEFEF}15.45 &\cellcolor[HTML]{EFEFEF}22.79 & 21.66 &\cellcolor[HTML]{EFEFEF}21.89 \\
DeOP & $-\mathcal{L}_\mathrm{DeOP}$ & 36.27 & 46.31 & 29.75 & 43.17 & 44.58 &\cellcolor[HTML]{EFEFEF}17.22 &\cellcolor[HTML]{EFEFEF}13.61 & 18.21 &\cellcolor[HTML]{EFEFEF} 19.28 \\
 \hline
 & DR & 52.81 & 58.01 & 46.77 & 53.14 & 55.95 &\cellcolor[HTML]{EFEFEF}{12.64} &\cellcolor[HTML]{EFEFEF}23.94 & 24.70 &\cellcolor[HTML]{EFEFEF}25.71\\
 & NRD & 57.51 & 60.87 & 35.03 & 47.77 & 49.99 &\cellcolor[HTML]{EFEFEF}15.69 &\cellcolor[HTML]{EFEFEF}15.59 & 21.50 &\cellcolor[HTML]{EFEFEF}22.00\\
\multirow{-3}{*}{\begin{tabular}[c]{@{}c@{}}CLIP\\ ViT-B\end{tabular}} & \textbf{PRM} & \bluebf{13.71} & \bluebf{26.11} & \bluebf{8.79} & \bluebf{22.71} & \bluebf{24.80} &\cellcolor[HTML]{EFEFEF}\bluebf{1.83} &\cellcolor[HTML]{EFEFEF}\bluebf{4.87} & \bluebf{11.22} &\cellcolor[HTML]{EFEFEF}\bluebf{11.32} \\
 \hline \hline
 & DR & 39.08 &\cellcolor[HTML]{EFEFEF}44.39 & 36.04 &  41.39 &\cellcolor[HTML]{EFEFEF}27.55 & 21.75 & 21.11 & 23.93 & 24.50\\
 & NRD & 35.33 &\cellcolor[HTML]{EFEFEF}40.22 & 32.33 & 38.14 &\cellcolor[HTML]{EFEFEF}22.60 & 21.46 & 20.29 & 23.68 & 24.19\\
\multirow{-3}{*}{\begin{tabular}[c]{@{}c@{}}CLIP\\ {CNeXt-L}\end{tabular}} & \textbf{PRM} & \bluebf{6.99} &\cellcolor[HTML]{EFEFEF}\bluebf{10.44} & \bluebf{5.97} & \bluebf{8.17} &\cellcolor[HTML]{EFEFEF}\bluebf{0.93} & \bluebf{12.64} & \bluebf{13.10} & \bluebf{18.94} & \bluebf{19.38}\\
\specialrule{.8pt}{1pt}{1pt}
\end{tabular}
\end{subtable}
\vspace{-15pt}
\end{table}

\begin{table}[t] 
\caption{Transfer attack efficacy on OVD target models. Model performance is measured in \textbf{AP50}. Metrics of Base (B) and novel (N) classes are reported separately.}\label{tab:ovd}
\centering\fontsize{6}{8}\selectfont
\begin{tabular}{cc|C{.7cm}C{.7cm}|C{.7cm}C{.7cm}|C{.7cm}C{.7cm}|C{.7cm}C{.7cm}|C{.7cm}C{.7cm}|C{.7cm}C{.7cm}}
\specialrule{.8pt}{1pt}{1pt}
 &  & \multicolumn{2}{c|}{Detic} & \multicolumn{2}{c|}{OV-DETR } & \multicolumn{2}{c|}{CORA } & \multicolumn{2}{c|}{CoDet } & \multicolumn{2}{c|}{OC-OVD} & \multicolumn{2}{c}{VL-PLM } \\
  \cline{3-14}
 &  & \multicolumn{2}{c|}{ViT-B/16} & \multicolumn{2}{c|}{ViT-B/16 } & \multicolumn{2}{c|}{RN50} & \multicolumn{2}{c|}{ViT-B/16 } & \multicolumn{2}{c|}{ViT-B/16 } & \multicolumn{2}{c}{ViT-B/16 } \\
  \cline{3-14}
 &  \multirow{-3}{*}{$\mathcal{L}_\mathrm{attack}$}  & B & N & B & N & B & N & B & N & B & N & B & N \\
  \cline{2-14}
\multirow{-4}{*}{\backslashbox[8mm]{S}{T}} 
& Clean & 51.10 & 27.81 & 55.67 & 30.00 & 36.81 & 35.49 & 52.47 & 30.63 & 56.58 & 40.38 & 56.37 & 34.07 \\
 \hline \hline
\rowcolor[HTML]{EFEFEF} 
\cellcolor[HTML]{FFFFFF}  SAN & \cellcolor[HTML]{FFFFFF}  $-\mathcal{L}_\mathrm{SAN}$ & 39.42 & 18.83 & 42.78 & 20.34 &\cellcolor[HTML]{FFFFFF}  26.31 & \cellcolor[HTML]{FFFFFF}  25.83 & 40.85 & 19.83 & 44.07 & 29.35 & 46.33 & 27.80 \\

\rowcolor[HTML]{EFEFEF} 
\cellcolor[HTML]{FFFFFF}   DeOP &\cellcolor[HTML]{FFFFFF}   $-\mathcal{L}_\mathrm{DeOP}$ & 38.65 & 16.48 & 45.00 & 21.58 &\cellcolor[HTML]{FFFFFF}   19.82 & \cellcolor[HTML]{FFFFFF}  18.00 & 39.69 & 17.91 & 42.66 & 26.94 & 46.98 & 27.25  \\
 \hline
& DR &\cellcolor[HTML]{EFEFEF}47.50 &\cellcolor[HTML]{EFEFEF}25.74 &\cellcolor[HTML]{EFEFEF}52.83 &\cellcolor[HTML]{EFEFEF}27.18 &\cellcolor[HTML]{FFFFFF}   31.94 &\cellcolor[HTML]{FFFFFF}   31.07 &\cellcolor[HTML]{EFEFEF}48.69 &\cellcolor[HTML]{EFEFEF}26.92 &\cellcolor[HTML]{EFEFEF}52.20 &\cellcolor[HTML]{EFEFEF}36.51 &\cellcolor[HTML]{EFEFEF}54.37 &\cellcolor[HTML]{EFEFEF}32.94 \\
& NRD &\cellcolor[HTML]{EFEFEF}40.75 &\cellcolor[HTML]{EFEFEF}19.58 &\cellcolor[HTML]{EFEFEF}45.67 &\cellcolor[HTML]{EFEFEF}45.67 & \cellcolor[HTML]{FFFFFF}  23.02 & \cellcolor[HTML]{FFFFFF}  22.06 &\cellcolor[HTML]{EFEFEF}41.98 &\cellcolor[HTML]{EFEFEF}20.05 &\cellcolor[HTML]{EFEFEF}44.39 &\cellcolor[HTML]{EFEFEF}28.15 &\cellcolor[HTML]{EFEFEF}49.05 &\cellcolor[HTML]{EFEFEF}28.84 \\
\multirow{-3}{*}{ \begin{tabular}[c]{@{}c@{}}  CLIP\\  ViT-B\end{tabular}} & \textbf{PRM} &\cellcolor[HTML]{EFEFEF}\bluebf{20.12} &\cellcolor[HTML]{EFEFEF}\bluebf{5.17} &\cellcolor[HTML]{EFEFEF}\bluebf{21.60} &\cellcolor[HTML]{EFEFEF}\bluebf{5.90} & \cellcolor[HTML]{FFFFFF}\bluebf{7.13} & \cellcolor[HTML]{FFFFFF}\bluebf{4.47} &\cellcolor[HTML]{EFEFEF}\bluebf{21.04} &\cellcolor[HTML]{EFEFEF}\bluebf{5.70}&\cellcolor[HTML]{EFEFEF}\bluebf{21.24}&\cellcolor[HTML]{EFEFEF}\bluebf{8.90}&\cellcolor[HTML]{EFEFEF}\bluebf{22.95}&\cellcolor[HTML]{EFEFEF}\bluebf{9.54} \\
 \hline \hline
 & DR & 38.89 & 17.31 & 46.01 & 22.24 & 20.97 & 18.79 & 39.87 & 18.39 & 43.52 & 27.74 & 48.61 & 29.22 \\
 & NRD & 36.75 & 15.41 & 44.26 & 20.71 & 19.03 & 16.09 & 37.97 & 16.16 & 40.87 & 24.96 & 46.93 & 26.98 \\
\multirow{-3}{*}{\begin{tabular}[c]{@{}c@{}}CLIP\\ {CNeXt-L}\end{tabular}} & \textbf{PRM} & \bluebf{15.37} &\bluebf{3.87} & \bluebf{17.12} & \bluebf{4.69} & \bluebf{3.39} & \bluebf{1.35} & \bluebf{15.61} & \bluebf{3.71} & \bluebf{15.99} & \bluebf{5.90} & \bluebf{22.01} & \bluebf{8.88}\\
\specialrule{.8pt}{1pt}{1pt}
\end{tabular}
\end{table}

\begin{table}[t]
\centering
\fontsize{6}{8}\selectfont
\caption{Transfer attack efficacy on image captioning and visual question-answering (VQA) target models. Caption quality is measured with BLEU score (B@1, B@4) \cite{cap-metric-papineni2002bleu}, METEOR (M) \cite{cap-metric-denkowski2014meteor}, ROUGE-L (R-L) \cite{cap-metric-lin2004rouge-l} and CIDEr (C) \cite{cap-metric-vedantam2015cider}. VQA performance is measured with VQA accuracy. ``VB'' and ``CL'' stands for  ViT-B/16 and ConvNeXt-L CLIP vision encoders as surrogates.}\label{tab:ic-vqa}
\begin{tabular}{C{0.4cm}c|ccccc|cccccc|cccccc}
\specialrule{.8pt}{1pt}{1pt}
 &  & \multicolumn{5}{c|}{CapDec (RN50$\times$4)} & \multicolumn{6}{c|}{OpenFlamingo (ViT-L/14) } &\multicolumn{6}{c}{LLaVA (ViT-L/14)}  \\
 \cline{3-19}
 & \multirow{-2}{*}{$\mathcal{L}$} & B@1 & B@4 & M & R-L & C & B@1 & B@4 & M & R-L & C & VQA & B@1 & B@4 & M & R-L & C & VQA\\
 \cline{2-19}
{\multirow{-3}{*}{\backslashbox[4mm]{S}{T}}} & Clean & 68.3 & 26.8 & 25.2 & 51.3 & 92.4 & 64.9 & 24.9 & 22.7 & 50.5 & 87.3 & 31.9 & 72.3 & 28.7 & 28.7 & 55.2 & 106.9 & 54.1 \\
\hline \hline

 \multicolumn{2}{c|}{SAN} &  64.2 & 23.6 & 23.3 & 48.7 & 82.3 &54.9 & 18.2 & 18.8 & 44.5 & 64.5 & 26.7 & 69.7 & 28.0 & 27.6 & 53.4 & 100.7 & 52.3\\
 \multicolumn{2}{c|}{DeOP} & 66.6 & 26.5 & 24.5 & 50.4 & 93.1 & 56.6 & 21.0 & 20.5 & 47.2 & 78.1 & 30.9 & 70.4 & 28.0 & 27.9 & 54.0 & 101.1 & 53.8 \\
 \hline
 & DR & 66.2 & 24.9 & 23.9 & 49.7 & 84.1 & 57.1 & 21.0 & 20.5 & 47.3 & 75.6 & 29.8 & 72.2 & 28.8 & 28.4 & 55.0 & 105.0 & 53.6 \\
 & NRD & 66.2 & 24.8 & 23.9 & 49.8 & 84.6 & 56.4 & 20.0 & 19.9 & 46.4 & 71.1& 28.3 & 71.1 & 27.7 & 27.7 & 54.0 & 101.1 & 52.6\\
\multirow{-3}{*}{\begin{tabular}[c]{@{}c@{}}V\\B\end{tabular}} & \textbf{PRM} & \bluebf{50.9} & \bluebf{13.8} & \bluebf{16.6} & \bluebf{39.5} & \bluebf{40.6} & \bluebf{51.4} & \bluebf{14.0} & \bluebf{15.9 }& \bluebf{39.9} & \bluebf{45.3} & \bluebf{20.8} & \bluebf{57.2} & \bluebf{16.8} & \bluebf{20.5} & \bluebf{43.6} & \bluebf{57.0}& \bluebf{37.6}\\
 \hline \hline
 & DR &  66.2 & 24.9 & 24.1 & 49.9 & 85.5 & 56.9 & 20.6 & 20.4 & 47.0 & 74.8& 29.7 & 71.9 & 28.4 & 28.2 & 54.7 & 104.4 & 53.7 \\
 & NRD &  65.5 & 24.6 & 23.9 & 49.6 & 83.3 & 56.9 & 20.6 & 20.5 & 47.0 & 74.3& 28.9 & 72.2 & 28.5 & 28.2 & 54.8 & 104.2 & 53.6\\
\multirow{-3}{*}{\begin{tabular}[c]{@{}c@{}}C\\ L\end{tabular}} & \textbf{PRM} & \bluebf{46.0} & \bluebf{10.6} & \bluebf{14.5} & \bluebf{36.3} & \bluebf{29.7} & \bluebf{56.4} & \bluebf{18.3} & \bluebf{18.8} & \bluebf{44.4} & \bluebf{63.5} & \bluebf{26.6}  & \bluebf{66.0} & \bluebf{23.5} & \bluebf{25.2} & \bluebf{50.0} & \bluebf{84.7} & \bluebf{50.2}\\
\specialrule{.8pt}{1pt}{1pt}
\end{tabular}
\vspace{-10pt}
\end{table}

\subsection{Cross-Task Attack Performance}
\label{subsec:main-results-discussions}

\noindent\textbf{Inheritance of Adversarial Vulnerability.}
As shown in Table \ref{tab:ovs-pc59}-\ref{tab:ic-vqa}, our attack transfers to a comprehensive collection of victim models with significantly different training paradigms across all four tasks of interest. Our method (PRM) produces significantly more effective adversaries in comparison to baseline methods with both surrogates, as highlighted by the blue numbers in the tables. This is also clearly illustrated by Figure \ref{fig:radial-plots} where the red line (PRM) shows a substantial shrinkage in performance across all target tasks and models.
This provides strong empirical evidence that downstream ``child'' models could inherit adversarial vulnerabilities from ``parent'' foundation models, rendering them more susceptible to transfer attacks created with the parent models.

% The phenomenon of such ``adversarial vulnerability inheritance'' has been observed in prior works in other transfer-learning contexts, \cite{bg-advdownstream-wang2018great} which have pointed out that deriving models from few publicly available teacher models could increase downstream student models' vulnerability to misclassification attacks devised with the teacher models. 
% % Since vision-language alignment pre-training is, to some extent, analogous to classification, it is expected that a similar form of vulnerability inheritance could occur in the case of CLIP.
% With our results, we can verify that such phenomenon indeed generalises to a wider range of transfer learning settings beyond those considered in the original study: (1) it generalises to tasks beyond classification; (2) it generalises to more impactful large foundation models and their downstream models, including scenarios where there is a significant model capacity gap between the pre-training and the downstream models; (3) it generalises to transfer-learning scenarios where the pre-training model and the downstream model have different learning objectives (i.e. unlike traditional settings where both pre-training and downstream models are trained with classification objective, CLIP is trained using a vision-language alignment objective whereas downstream models have diverse training paradigm). 

\noindent\textbf{Transferability Across Vision Encoder Variants.}
Despite prior research showing limited transferability of adversarial examples between models with significantly different parameters and architectures \cite{bg-advseg-arnab2018robustness,methods-motivation-waseda2023closer}, we notice that
% variations in vision encoder model parameters and architectures do not prevent adversarial samples from transferring across downstream models. In particular,
% (1) Perturbations that can cause feature distortion in the off-the-self version of CLIP vision encoder features can fool downstream models that fine-tunes their encoder backbones using various strategies, including full-body fine-tuning (OV-Seg\cite{bg-ovs-dense-liang2023ov-seg} and ZSSeg-Baesline \cite{bg-ovs-dense-xu2022zssegbaseline}) or finetuning of specific components such as positional embeddings (SAN\cite{bg-ovs-dense-xu2023san}, DeOP\cite{bg-ovs-dense-Han2023DeOP}) or attention modules (CAT-Seg\cite{bg-ovs-dense-cho2023catseg}). (2) 
substantial transferability can be attained even when a surrogate has a vision encoder architecture that does not match that of the target model. White cells in the tables indicate such scenarios, exemplified by the last three rows in the first half of Table \ref{tab:ovs-pc59}. For some target models, a mismatched surrogate model can even produce stronger attacks than a matched surrogate: ZSSeg-BL\cite{bg-ovs-dense-xu2022zssegbaseline} and ZegFormer\cite{bg-ovs-dense-ding2021dZegFormer} (columns 4-5, Table \ref{tab:ovs-pc59}) use ViT-B backbones but adversaries created with ConvNeXt-L cause more harm to them. Notably, none of the VQA and captioning target models share the same backbone as either surrogate yet our attack still induces significant performance degradation, as highlighted by the blue numbers in Table \ref{tab:ic-vqa}.
% There is also a particularly interesting observation among OVD victim models in Table \ref{tab:ovd}: all OVD target models have either ViT-B/16 or RN50 encoder backbones, yet the most effective attacks for all of them are crafted with ConvNeXt-L encoders.
% While prior works show that adversarial examples have limited transferability between models of distinct architectures (i.e. between CNN and ViT), this limitation does not seem to hold true in our case. 
% These observations imply that mitigating downstream models' susceptibility to transfer-based attacks due to these common vulnerabilities is a non-trivial task: they cannot be naively alleviated by avoiding shared parameters through task-specific fine-tuning or using an encoder with a different architecture than the one available to an attacker.
These observations suggest a significant overlap of non-robust features \cite{methods-motivation-waseda2023closer} across vision encoders of various architectures, which is likely due to the common vision-language alignment pre-training process.

\noindent\textbf{Transferability Across Tasks.}
Interestingly, as exemplified by Figure \ref{fig:qualitative-2008_000788}, PRM perturbations often incur semantically consistent mistakes across downstream models of various tasks (e.g. in this sample, most models made mistakes by hallucinating a false positive human in this scene). Closer scrutiny reveals that certain perturbations on their own may exhibit semantic meanings \cite{apx-semantic-attack-ilyas2019adversarial,apx-semantic-attack-tao2018attacks}: comparing the LLaVA captions of the clean image $x$, the adversarial sample $x'$ and the perturbation $\delta$, they seem to exhibit an additive relationship in terms of perceived semantic meanings.

\definecolor{brightred}{HTML}{FF9999}
\definecolor{brightgreen}{HTML}{97D077}
\tikzset{every picture/.style={/utils/exec={\sffamily}}}

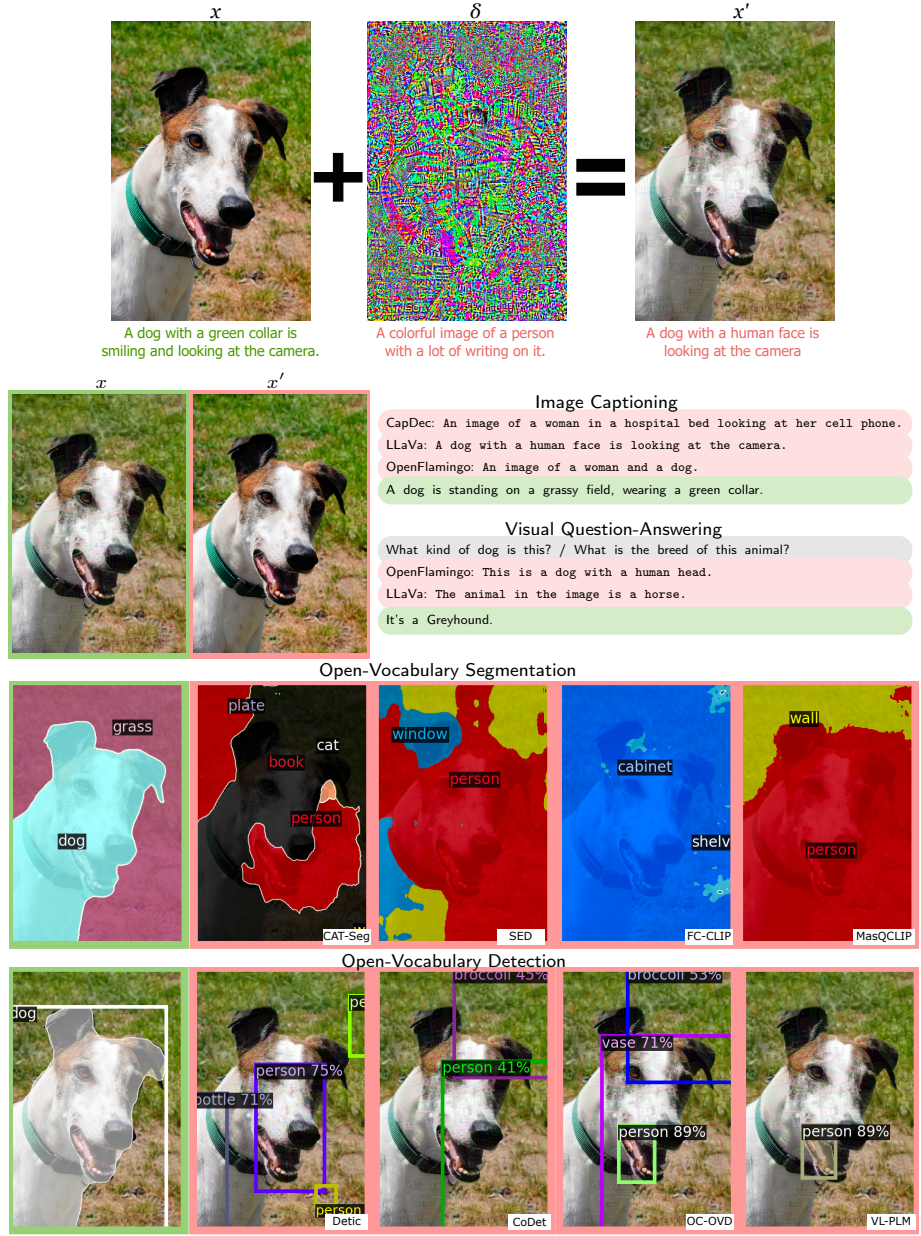
\begin{figure}[H]
\centering
\includegraphics[width=.8\textwidth]{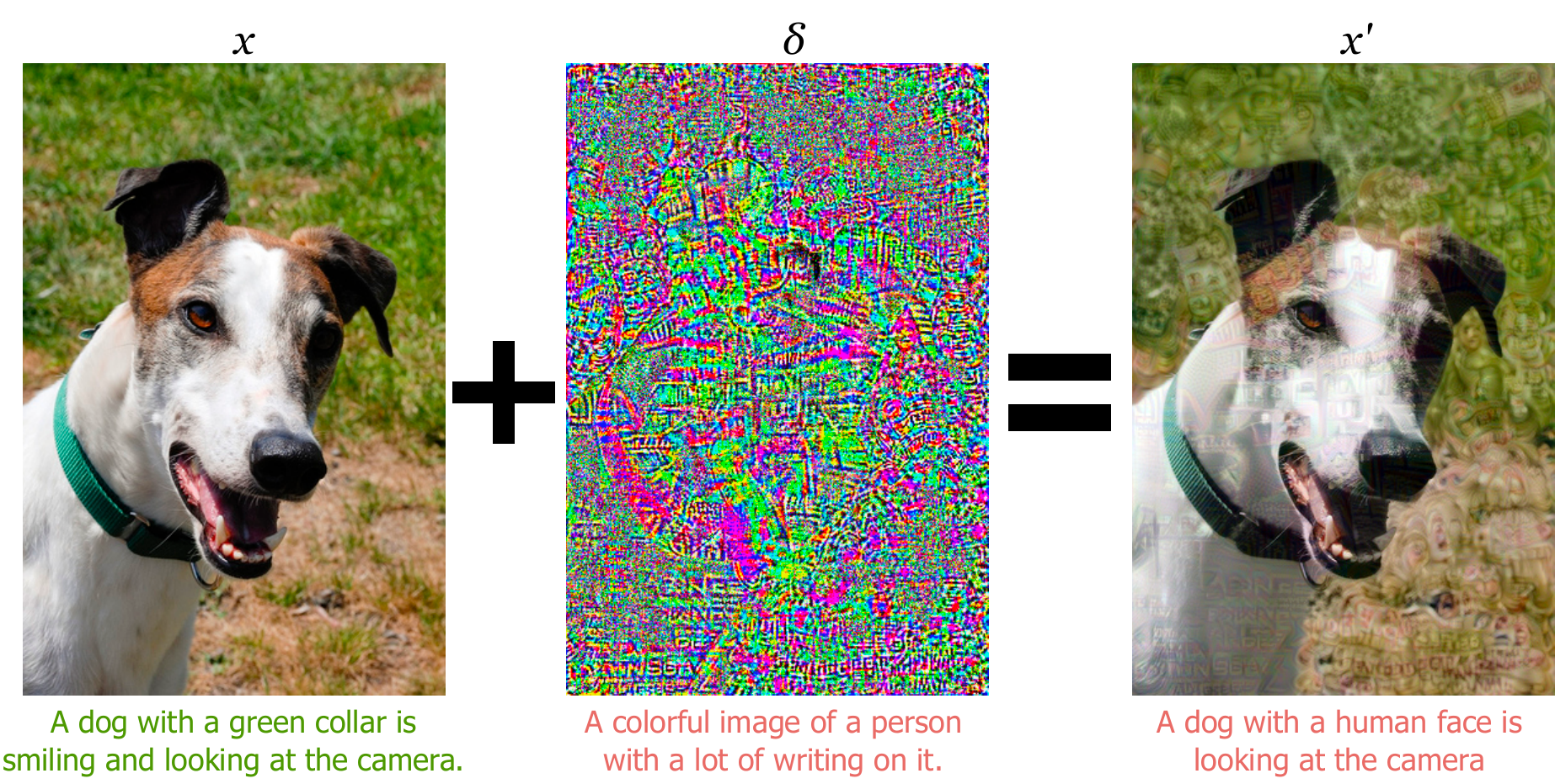}\\
\resizebox{\columnwidth}{!}{%
\begin{tikzpicture}[]

    \fill [fill=brightgreen,anchor=north west,inner sep=0] (2.4,6.75) rectangle (-0.05,3.15);
    \fill [fill=brightred,anchor=north west,inner sep=0] (4.8,6.75) rectangle (2.38,3.15);
    % \fill [fill=brightred!30,anchor=north west,inner sep=0] (12.1,3.3) rectangle (4.8,-3.9);
    % \fill [fill=white,anchor=north west,inner sep=0] (-0.1,3) rectangle (12.1,3.2);
    % \fill [fill=white,anchor=north west,inner sep=0] (-0.1,-.4) rectangle (12.1,-.2);
    
    % \node [anchor=north west,rotate=0]at (1,4.65) {\tiny SAN};
    % \node [anchor=north west,rotate=0]at (3.7,4.65) {\tiny DeOP};
    % \node [anchor=north west,rotate=0]at (6.5,4.65) {\tiny FC-CLIP};
    % \node [anchor=north west,rotate=0]at (9.2,4.65) {\tiny CAT-Seg-L};
    \node[anchor=north west,inner sep=0] at (0,6.7){
    \includegraphics[width=0.19\linewidth,trim={0 0 0 0cm},clip]{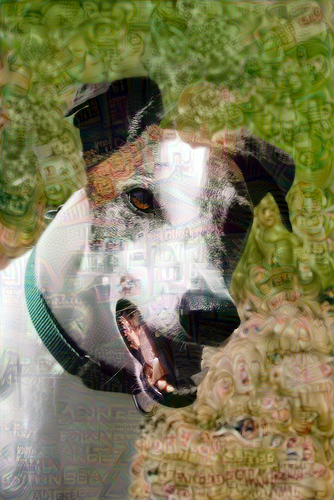}
    \includegraphics[width=0.19\linewidth,trim={0 0 0 0cm},clip]{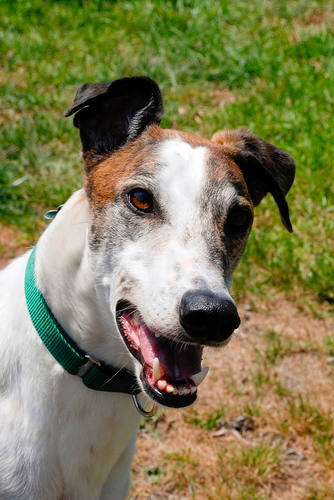}
    };
    
    \node[anchor=north west,inner sep=0] at (-0.05,2.85) {
    \includegraphics[width=\textwidth]{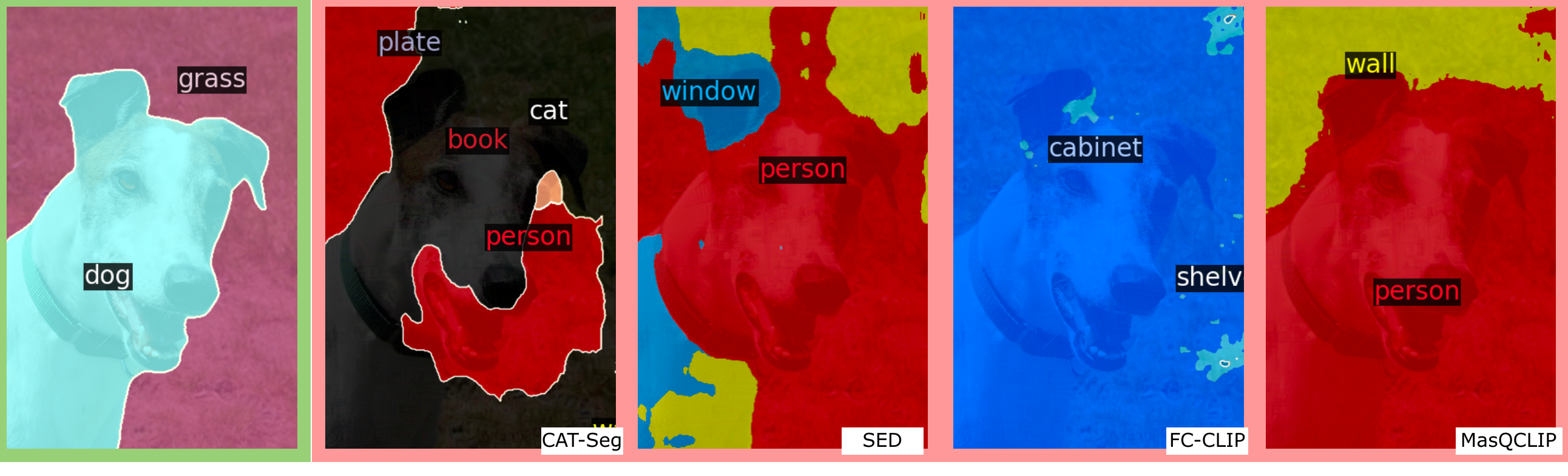}
    }; 
    
    \node[anchor=north west,inner sep=0] at (-0.05,-1.0) {
    \includegraphics[width=\textwidth]{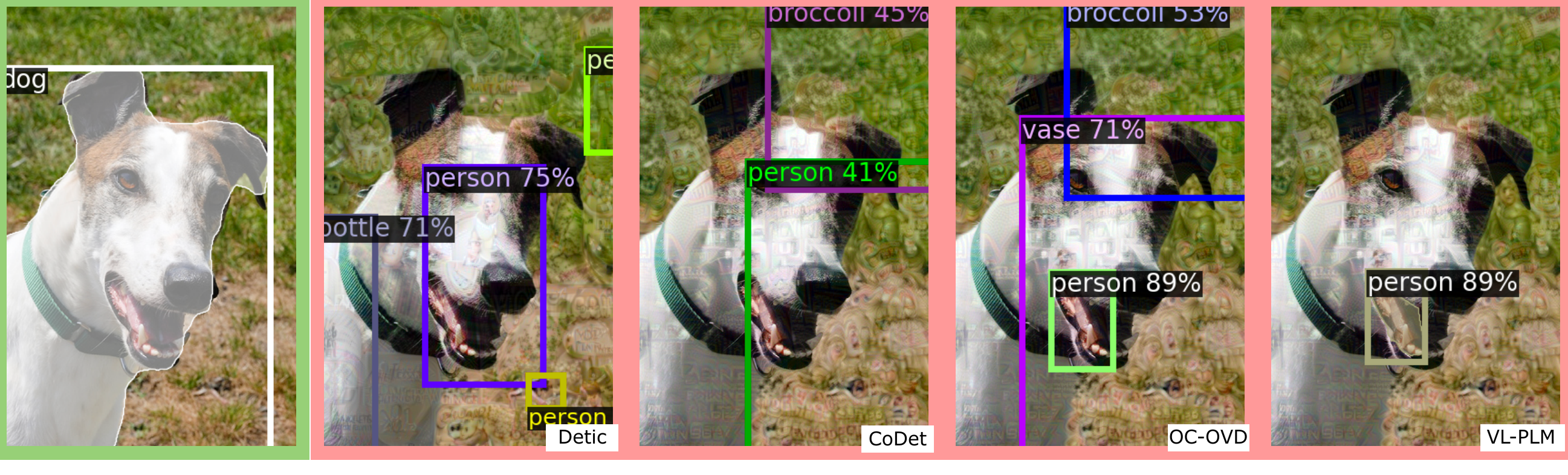}
    }; 
    
    % \node [anchor=north west,rotate=0] at (0,-4.15) [fill, rounded corners=5pt, fill=bubblegrey!20,] {\tiny Please describe this image using one sentence.};
    \node [anchor=north west,rotate=0,align=left, text width=0.57\textwidth] at (4.9,6.5) [fill, rounded corners=5pt, fill=brightred!30,] {\tiny CapDec: \texttt{An image of a woman in a hospital bed looking at her cell phone.}};
    \node [anchor=north west,rotate=0,align=left, text width=0.57\textwidth] at (4.9,6.2)[fill, rounded corners=5pt, fill=brightred!30,] {\tiny LLaVa: \texttt{A dog with a human face is looking at the camera.}};
    \node [anchor=north west,rotate=0,align=left, text width=0.57\textwidth] at (4.9,5.9)[fill, rounded corners=5pt, fill=brightred!30,] {\tiny OpenFlamingo: \texttt{An image of a woman and a dog.}};
    \node [anchor=north west,rotate=0,align=left, text width=0.57\textwidth] at (4.9,5.6) [fill, rounded corners=5pt, fill=brightgreen!40,] {\tiny A dog is standing on a grassy field, wearing a green collar.};

    \node [anchor=north west,rotate=0,align=left, text width=0.57\textwidth] at (4.9,4.8) [fill, rounded corners=5pt, fill=gray!20, text width=0.57\textwidth] {\tiny What kind of dog is this? / What is the breed of this animal?};
    \node [anchor=north west,rotate=0,align=left, text width=0.57\textwidth] at (4.9,4.5) [fill, rounded corners=5pt, fill=brightred!30,] {\baselineskip=5pt \tiny OpenFlamingo:  \texttt{This is a dog with a human head.} \par};
    \node [anchor=north west,rotate=0,align=left, text width=0.57\textwidth] at (4.9,4.2) [fill, rounded corners=5pt, fill=brightred!30,]  {\baselineskip=5pt \tiny LLaVa: \texttt{The animal in the image is a horse.} \par};
    \node [anchor=north west,rotate=0,align=left, text width=0.57\textwidth] at (4.9,3.85) [fill, rounded corners=5pt, fill=brightgreen!40,] {\baselineskip=5pt \tiny It's a Greyhound.\par};

    \node [anchor=north west,rotate=0]at (4,3.2) {\scriptsize Open-Vocabulary Segmentation};
    \node [anchor=north west,rotate=0]at (4.3,-0.7) {\scriptsize Open-Vocabulary Detection};
    \node [anchor=north west,rotate=0]at (6.9,6.8) {\scriptsize Image Captioning};
    \node [anchor=north west,rotate=0]at (6.5,5.1) {\scriptsize Visual Question-Answering};
    \node [anchor=north west,rotate=0]at (1. ,7) {\scriptsize $x$};
    \node [anchor=north west,rotate=0]at (3.3,7.1) {\scriptsize $x'$};
\end{tikzpicture}
}
\caption{
An adversarial example created via PRM (using ViT-B/16 CLIP vision encoder as a surrogate) can fool various downstream models across various tasks. Clean ($x$) and adversarial ($x'$) inputs are shown in the top-left corner. Correct predictions are marked with green frames whereas adversarial predictions are marked with red. Note that downstream models tend to make semantically consistent mistakes (i.e. perceiving a false positive human in the scene).}
\label{fig:qualitative-2008_000788}
\end{figure}

\noindent While significant performance degradation is observed on all four tasks of interest, our attack demonstrates particularly strong efficacy among dense predictors (OVS and OVD models in Tables \ref{tab:ovs-pc59}-\ref{tab:ovd}), where the impact of full-coverage dense semantic distortion induced by our patch-wise cosine similarity minimisation objective is more prominent.
We also point out that OVD target models' performance on novel classes that are unseen during training (``N'' columns in Table \ref{tab:ovd}) suffer more performance degradation (our best attack yields an average adversarial performance equivalent to 29\% of the clean performance on base classes and 14\% of the clean performance on novel classes). As the models' generalisability to unseen classes relies heavily on the pre-trained cross-modality semantics of CLIP, it is expected that the disruption to pre-trained features impacts their performance on unseen classes to a larger extent.
On the other hand, we notice that all strategies presented in this study have a relatively small impact on VQA target models: PRM is the only method that achieves substantial attack efficacy with an average adversarial performance equivalent to 67\% of the clean performance while other methods generally cannot degrade VQA accuracy by more than 3\%.
This may be attributed to the fact that VQA models have a tendency to learn \textit{linguistic shortcuts} \cite{vqa-ling-shortcut-chen2020counterfactual,vqa-ling-shortcut-goyal2017making,vqa-ling-shortcut-jabri2016revisiting,vqa-ling-shortcut-niu2021counterfactual,vqa-multi-shortcut-dancette2021beyond,vqa-multi-shortcut-si2022language}, a form of bias and spurious correlation in the language modality that reduces their reliance on pre-trained vision features.
Developing a stronger attack for tasks that have a greater emphasis on language (e.g. by involving the language modality in attack optimisation \cite{transfer-prompt-luo2023image} or explicitly leveraging contextual hints such as class co-occurrence information \cite{bg-adv-transfer-aich2022gama}) could be an interesting direction for future work.

% This may be attributed to the fact that VQA models have a tendency to learn \textit{linguistic shortcuts} (linguistic\cite{vqa-ling-shortcut-chen2020counterfactual,vqa-ling-shortcut-goyal2017making,vqa-ling-shortcut-jabri2016revisiting,vqa-ling-shortcut-niu2021counterfactual} or multimodal \cite{vqa-multi-shortcut-dancette2021beyond,vqa-multi-shortcut-si2022language}), a form of bias and spurious correlation that has strong association with the language modality and thus not accounted for by attack strategies that involve solely the vision modality. Developing a stronger attack that considers these shortcuts (e.g. by involving language and/or explicitly leveraging contextual hints such as class co-occurrence matrices) could be an interesting direction for future work.

\subsection{Discussions and Ablations}\label{subsec:ablations}
\subsubsection{Cosine Similarity-Based Distortion.}\label{subsec:ablation-cosine-dim}

We hypothesised two potential factors that may contribute to the efficacy of our loss objective: (1) a better ability to induce dense semantic perturbations across all image patches; (2) the choice of angular distance as a more appropriate distance metric in the CLIP dense vision feature space. 
In the main experiments, we compare our patch-wise cosine similarity-based approach with global MSE-based (NRD) \cite{bg-transfer-crosstask-naseer2018task} or variance-based (DR) \cite{bg-transfer-crosstask-lu2020dispersion-reduction} feature distortion losses adopted by prior works and empirically show that our objective achieves better attack transferability. We examine our loss design in two alternative settings to further validate its efficacy.

\vspace{-5pt}\paragraph{Does alignment pretraining prioritize directional information to a greater extent?}
We repeat our experiments using architecturally identical non-CLIP vision encoders trained with classification objectives as surrogates. As shown in Table \ref{tab:cosine-nonclip} (Appendix \nolink{\ref{apx:subsec:flatten}}), while PRM still outperforms baselines, their performance gap shrinks notably (e.g. on selected OVS victim models, the performance gap between PRM and MSE-based NRD shrinks from 23\% on CLIP vision encoders to around 11\% on classification-trained encoders). This suggests that vision-language alignment pre-training could have played a role in assigning greater semantic importance to directional information over magnitude or variance (in comparison to classification cross entropy-trained alternatives).
% While the misalignment objective enjoys a larger performance gain when applied to surrogates pre-trained with a vision-language alignment objective due to its emphasis on angular distance, it can still be more effective than baselines on other encoders due to its capability to induce dense perturbations.

% cosine guarantees dense perturbation (each spatial element will surely be perturbed
% Empirically, this hypothesis is supported by the fact that our approach outperforms  (distorting directional information causes more harm than distorting feature magnitude or variance).
%  may regions of images with features that exhibit certain numerical traits (i.e. the absolute magnitudes or variances of network activation).
\vspace{-5pt}
\paragraph{Dimensions along which to compute cosine similarity.}
% \paragraph{The patch-wise approach outperforms the global approach.}
We attempt an alternative usage of cosine similarity that flattens features before cosine similarity computation (i.e. globally distorting the entire volumetric feature using cosine similarity). As shown in \nolink{Table \ref{tab:ablation-flatten} (Appendix \ref{apx:subsec:flatten}}), while this configuration still outperforms MSE-based and variance-based baselines, it is not as performant as our default patch-wise setup which treats every patch descriptor as an independent sample and computes cosine similarity specifically along the embedding (ViT) or channel (CNN) dimensions of the features. 
This drop in performance indicates that the patch-wise approach tends to be more effective than the global flattened approach even when both are similarly based on cosine similarity.

\vspace{-10pt}
\subsubsection{Which Features to Attack? }\label{subsec:ablation-layers}
As mentioned in Sec. \ref{subsec:setup}, we choose to attack all tokens (or patches) from all vision encoder layers. This contrasts with previous studies that argue single-layer attacks (only distorting features from one surrogate layer) on mid-layer features would be sufficient \cite{bg-transfer-crosstask-naseer2018task,bg-transfer-crosstask-lu2020dispersion-reduction,bg-transfer-feature-salzmann2021learning}. To determine whether attacking all features is necessary, we attempt several alternatives that attack subsets of features and present results in \nolink{Table \ref{tab:ablation-whichfeatures}, Appendix \ref{apx:subsec:whichfeatures}.}
% \begin{enumerate}
%     \item Only attacking \texttt{CLS} tokens from all layers (ViT-B surrogate). 
%     \item Only attacking the final layer features (ViT-B/ConvNeXt-L surrogates). 
%     \item Only attacking mid-layer features (ViT-B/ConvNeXt-L surrogates). 
% \end{enumerate}
We found that attacking all layers of the model is more effective than just attacking the final encoder layer (which directly affects model decisions) or attacking one intermediate layer (as recommended by prior works).

% We note that both the intuitive choice of attacking the final encoder layer  and the mid-layer attack recommended by prior works are not as effective as an all-layer attack. Moreover, local features are 
% Some prior works suggest that different neural networks tend to share commonalities in their mid-layer features \cite{bg-transfer-feature-salzmann2021learning} and designed attack strategies that perturb mid-level features. Other works empirically find the surrogate layer that leads to best transferability (e.g. conv3-3 of VGG16 as per \cite{bg-transfer-crosstask-lu2020dispersion-reduction}) and base their attack on that layer. 

\vspace{-10pt}
\subsubsection{The Effect of Input Scale Diversity.}\label{subsec:ablation-scale}
In the main experiments, we employ random rescaling augmentation for all competing loss objectives. We ablate this component and show results without input scale augmentation in \nolink{Table \ref{tab:ablation-scale}, Appendix \ref{apx:subsec:scale}}. 
% This technique significantly amplifies attack efficacy for all attacks by fostering scale invariance, which makes them more adept at fooling downstream models that operate at different input resolutions. 
While this technique significantly amplifies attack efficacy for all attacks by fostering scale invariance, our strategy still outperforms baselines when all are examined in the absence of such augmentations. 
% While augmentation is usually considered orthogonal to loss objective design, we do notice that input diversity magnifies the performance gap between our method and the baselines.

\vspace{-10pt}
\subsubsection{The Role of Pre-Training.}\label{subsec:alignment-pretraining}
% The fact that attacks crafted with either ViT or ConvNeXt surrogate can transfer to downstream models regardless of the target model's vision encoder architecture (e.g. that use a different vision encoder backbone, as discussed in Sec. \ref{subsec:main-results-discussions}) prompts us to consider how certain factors in the pre-training process may contribute to similarities in adversarial vulnerabilities across CLIP's downstream models. 
% While the root cause of adversarial transferability remains an open question, there is strong evidence that spurious correlations picked up by the contribute to common vulnerabilities to input perturbations.
Various factors in the pre-training process (e.g. data, training objective and optimisation techniques) can contribute to the generalisability of an encoder's feature space. To examine how different pretraining protocols affect an encoder's efficacy as an attack surrogate, we perform our attack using architecturally identical vision encoders pre-trained and fine-tuned with different protocols (\nolink{Table \ref{tab:ablation-alignment-pretraining}, Appendix \ref{apx:subsec:alignment-pretraining}}). Notably, adversaries crafted with classification-trained encoders exhibit much lower transferability compared to those crafted with a vision-language alignment-pretrained vision encoder of the same architecture, suggesting that the latter encoder serves as a better tool for identifying common non-robust features on which downstream models rely.
% We hypothesise that the combination of web-scale vision-language data, the alignment training objective and various training configurations make alignment-trained encoders better feature extractors and hence the basis of more transferable attacks.
% This implies that vision-language alignment pre-trained encoders can be used to identify more common non-robust features on which downstream models rely.
% However, we note that this set of experiments is no grounds for concluding that the difference in the pre-training objective is the sole factor contributing to the encoders becoming more effective as attack surrogates since other factors in the pre-training process (e.g. data and optimisation techniques) could also contribute to this difference in their efficacy as attack surrogates - a more thorough discussion can be found in Appendix \ref{apx:subsec:alignment-pretraining}.

\section{Conclusions}
This work aims to raise awareness about a currently overlooked but widespread safety vulnerability introduced by the usage of open-sourced foundation models, such as CLIP, to its downstream applications.
% Pre-training not only provides valuable knowledge for task learning but also creates grounds for potential adversarial attacks. 
We show that the downstream models' reliance on pre-trained features can be exploited by attackers, who may devise extremely effective adversarial attacks using solely open-sourced foundation models. Through the design of a highly transferable cross-task adversarial attack strategy termed Patch Representation Misalignment (PRM), we demonstrate that such feature-based attacks can significantly compromise the performance of the downstream models independently of the type of task or architecture they adopt. We remark on the extent of such vulnerabilities by showcasing the effectiveness of our PRM attack on 20 different models across 4 common vision-language tasks. 
% The alarming efficacy of PRM raises concerns about the safety risk introduced by the current widespread practice of reusing foundation model components in order to develop systems that are becoming increasingly pervasive across industrial applications. 
While our investigation focuses on CLIP's downstream systems, the alarming efficacy of PRM warrants additional efforts to determine whether similar phenomena apply to other foundational models. We hope our work will encourage further exploration into the safety implications of using foundation models in downstream learning, as well as inspire the development of effective defence strategies or robust learning approaches.

\bibliographystyle{splncs04}
\bibliography{main}

\begin{thebibliography}{100}
\providecommand{\url}[1]{\texttt{#1}}
\providecommand{\urlprefix}{URL }
\providecommand{\doi}[1]{https://doi.org/#1}

\bibitem{bg-ovs-dense-Han2023DeOP}
Zero-Shot Semantic Segmentation with Decoupled One-Pass Network (2023)

\bibitem{method-advseg-agnihotri2023cospgd}
Agnihotri, S., Keuper, M.: Cospgd: a unified white-box adversarial attack for pixel-wise prediction tasks. arXiv preprint arXiv:2302.02213  (2023)

\bibitem{bg-adv-transfer-aich2022gama}
Aich, A., Ta, C.K., Gupta, A.A., Song, C., Krishnamurthy, S., Asif, M.S., Roy-Chowdhury, A.: {GAMA}: Generative adversarial multi-object scene attacks. In: Thirty-Sixth Conference on Neural Information Processing Systems (2022), \url{https://openreview.net/forum?id=DRckHIGk8qw}

\bibitem{bg-advseg-arnab2018robustness}
Arnab, A., Miksik, O., Torr, P.H.: On the robustness of semantic segmentation models to adversarial attacks. In: Proceedings of the IEEE conference on computer vision and pattern recognition. pp. 888--897 (2018)

\bibitem{bg-ic-vqa-awadalla2023openflamingo}
Awadalla, A., Gao, I., Gardner, J., Hessel, J., Hanafy, Y., Zhu, W., Marathe, K., Bitton, Y., Gadre, S., Sagawa, S., et~al.: Openflamingo: An open-source framework for training large autoregressive vision-language models. arXiv preprint arXiv:2308.01390  (2023)

\bibitem{dataset-caesar2018coco-stuff}
Caesar, H., Uijlings, J., Ferrari, V.: Coco-stuff: Thing and stuff classes in context. In: Proceedings of the IEEE conference on computer vision and pattern recognition. pp. 1209--1218 (2018)

\bibitem{bg-foundationsafety-adv-carlini2023poisoning}
Carlini, N., Jagielski, M., Choquette-Choo, C.A., Paleka, D., Pearce, W., Anderson, H., Terzis, A., Thomas, K., Tram{\`e}r, F.: Poisoning web-scale training datasets is practical. arXiv preprint arXiv:2302.10149  (2023)

\bibitem{bg-ovs-sparse-cha2022tcl}
Cha, J., Mun, J., Roh, B.: Learning to generate text-grounded mask for open-world semantic segmentation from only image-text pairs. In: Proceedings of the IEEE/CVF Conference on Computer Vision and Pattern Recognition (CVPR) (2023)

\bibitem{bg-transfer-surrogate-Chen_2023_ICCV}
Chen, B., Yin, J., Chen, S., Chen, B., Liu, X.: An adaptive model ensemble adversarial attack for boosting adversarial transferability. In: Proceedings of the IEEE/CVF International Conference on Computer Vision (ICCV). pp. 4489--4498 (October 2023)

\bibitem{bg-advdownstream-chen2024catastrophic}
Chen, H., Raj, B., Xie, X., Wang, J.: On catastrophic inheritance of large foundation models. arXiv preprint arXiv:2402.01909  (2024)

\bibitem{bg-transfer-generative-chen2023DiffAttack}
Chen, J., Chen, H., Chen, K., Zhang, Y., Zou, Z., Shi, Z.: Diffusion models for imperceptible and transferable adversarial attack. arXiv preprint arXiv:2305.08192  (2023)

\bibitem{vqa-ling-shortcut-chen2020counterfactual}
Chen, L., Yan, X., Xiao, J., Zhang, H., Pu, S., Zhuang, Y.: Counterfactual samples synthesizing for robust visual question answering. In: Proceedings of the IEEE/CVF conference on computer vision and pattern recognition. pp. 10800--10809 (2020)

\bibitem{chen2015coco-caption}
Chen, X., Fang, H., Lin, T.Y., Vedantam, R., Gupta, S., Doll{\'a}r, P., Zitnick, C.L.: Microsoft coco captions: Data collection and evaluation server. arXiv preprint arXiv:1504.00325  (2015)

\bibitem{chen2020uniter}
Chen, Y.C., Li, L., Yu, L., El~Kholy, A., Ahmed, F., Gan, Z., Cheng, Y., Liu, J.: Uniter: Universal image-text representation learning. In: European conference on computer vision. pp. 104--120. Springer (2020)

\bibitem{bg-ovs-dense-cho2023catseg}
Cho, S., Shin, H., Hong, S., An, S., Lee, S., Arnab, A., Seo, P.H., Kim, S.: Cat-seg: Cost aggregation for open-vocabulary semantic segmentation (2023)

\bibitem{vqa-multi-shortcut-dancette2021beyond}
Dancette, C., Cadene, R., Teney, D., Cord, M.: Beyond question-based biases: Assessing multimodal shortcut learning in visual question answering. In: Proceedings of the IEEE/CVF International Conference on Computer Vision. pp. 1574--1583 (2021)

\bibitem{cap-metric-denkowski2014meteor}
Denkowski, M., Lavie, A.: Meteor universal: Language specific translation evaluation for any target language. In: Proceedings of the ninth workshop on statistical machine translation. pp. 376--380 (2014)

\bibitem{bg-ovs-dense-ding2021dZegFormer}
Ding, J., Xue, N., Xia, G.S., Dai, D.: Decoupling zero-shot semantic segmentation  (2022)

\bibitem{bg-foundationsafety-dong2023robust}
Dong, Y., Chen, H., Chen, J., Fang, Z., Yang, X., Zhang, Y., Tian, Y., Su, H., Zhu, J.: How robust is google's bard to adversarial image attacks? arXiv preprint arXiv:2309.11751  (2023)

\bibitem{bg-transfer-aug-TIM}
Dong, Y., Pang, T., Su, H., Zhu, J.: Evading defenses to transferable adversarial examples by translation-invariant attacks. In: Proceedings of the IEEE/CVF Conference on Computer Vision and Pattern Recognition. pp. 4312--4321 (2019)

\bibitem{dosovitskiy2020vit}
Dosovitskiy, A., Beyer, L., Kolesnikov, A., Weissenborn, D., Zhai, X., Unterthiner, T., Dehghani, M., Minderer, M., Heigold, G., Gelly, S., et~al.: An image is worth 16x16 words: Transformers for image recognition at scale. arXiv preprint arXiv:2010.11929  (2020)

\bibitem{bg-ic-fei-iccv23-ViECap}
Fei, J., Wang, T., Zhang, J., He, Z., Wang, C., Zheng, F.: Transferable decoding with visual entities for zero-shot image captioning. In: Proceedings of the IEEE/CVF International Conference on Computer Vision (ICCV). pp. 3136--3146 (October 2023)

\bibitem{bg-transfer-feature-ganeshan2019fda}
Ganeshan, A., BS, V., Babu, R.V.: Fda: Feature disruptive attack. In: Proceedings of the IEEE/CVF International Conference on Computer Vision. pp. 8069--8079 (2019)

\bibitem{bg-adv-goodfellow2014explaining}
Goodfellow, I.J., Shlens, J., Szegedy, C.: Explaining and harnessing adversarial examples. arXiv preprint arXiv:1412.6572  (2014)

\bibitem{vqa-ling-shortcut-goyal2017making}
Goyal, Y., Khot, T., Summers-Stay, D., Batra, D., Parikh, D.: Making the v in vqa matter: Elevating the role of image understanding in visual question answering. In: Proceedings of the IEEE conference on computer vision and pattern recognition. pp. 6904--6913 (2017)

\bibitem{bg-transfer-gu2023survey}
Gu, J., Jia, X., de~Jorge, P., Yu, W., Liu, X., Ma, A., Xun, Y., Hu, A., Khakzar, A., Li, Z., et~al.: A survey on transferability of adversarial examples across deep neural networks. arXiv preprint arXiv:2310.17626  (2023)

\bibitem{method-advseg-gu2022segpgd}
Gu, J., Zhao, H., Tresp, V., Torr, P.H.: Segpgd: An effective and efficient adversarial attack for evaluating and boosting segmentation robustness. In: European Conference on Computer Vision. pp. 308--325. Springer (2022)

\bibitem{he2016resnet}
He, K., Zhang, X., Ren, S., Sun, J.: Deep residual learning for image recognition. In: Proceedings of the IEEE conference on computer vision and pattern recognition. pp. 770--778 (2016)

\bibitem{bg-advdownstream-hua2023initialization}
Hua, A., Gu, J., Xue, Z., Carlini, N., Wong, E., Qin, Y.: Initialization matters for adversarial transfer learning. arXiv preprint arXiv:2312.05716  (2023)

\bibitem{bg-transfer-surrogate-huang2023tsea}
Huang, H., Chen, Z., Chen, H., Wang, Y., Zhang, K.: T-sea: Transfer-based self-ensemble attack on object detection. In: Proceedings of the IEEE/CVF Conference on Computer Vision and Pattern Recognition. pp. 20514--20523 (2023)

\bibitem{bg-transfer-feature-huang2022defeat}
Huang, L., Gao, C., Liu, N.: Defeat: Decoupled feature attack across deep neural networks. Neural Networks  \textbf{156},  13--28 (2022)

\bibitem{bg-transfer-feature-huang2019enhancing}
Huang, Q., Katsman, I., He, H., Gu, Z., Belongie, S., Lim, S.N.: Enhancing adversarial example transferability with an intermediate level attack. In: Proceedings of the IEEE/CVF international conference on computer vision. pp. 4733--4742 (2019)

\bibitem{apx-semantic-attack-ilyas2019adversarial}
Ilyas, A., Santurkar, S., Tsipras, D., Engstrom, L., Tran, B., Madry, A.: Adversarial examples are not bugs, they are features. Advances in neural information processing systems  \textbf{32} (2019)

\bibitem{bg-transfer-feature-inkawhich2020perturbing}
Inkawhich, N., Liang, K., Wang, B., Inkawhich, M., Carin, L., Chen, Y.: Perturbing across the feature hierarchy to improve standard and strict blackbox attack transferability. Advances in Neural Information Processing Systems  \textbf{33},  20791--20801 (2020)

\bibitem{vqa-ling-shortcut-jabri2016revisiting}
Jabri, A., Joulin, A., Van Der~Maaten, L.: Revisiting visual question answering baselines. In: European conference on computer vision. pp. 727--739. Springer (2016)

\bibitem{jia2021ALIGN}
Jia, C., Yang, Y., Xia, Y., Chen, Y.T., Parekh, Z., Pham, H., Le, Q., Sung, Y.H., Li, Z., Duerig, T.: Scaling up visual and vision-language representation learning with noisy text supervision. In: International conference on machine learning. pp. 4904--4916. PMLR (2021)

\bibitem{johnson2016perceptual}
Johnson, J., Alahi, A., Fei-Fei, L.: Perceptual losses for real-time style transfer and super-resolution. In: Computer Vision--ECCV 2016: 14th European Conference, Amsterdam, The Netherlands, October 11-14, 2016, Proceedings, Part II 14. pp. 694--711. Springer (2016)

\bibitem{li2019visualbert}
Li, L.H., Yatskar, M., Yin, D., Hsieh, C.J., Chang, K.W.: Visualbert: A simple and performant baseline for vision and language. arXiv preprint arXiv:1908.03557  (2019)

\bibitem{bg-ic-li2023decap}
Li, W., Zhu, L., Wen, L., Yang, Y.: Decap: Decoding clip latents for zero-shot captioning via text-only training. arXiv preprint arXiv:2303.03032  (2023)

\bibitem{bg-ovs-trainingfree-li2023clipsurgery}
Li, Y., Wang, H., Duan, Y., Li, X.: Clip surgery for better explainability with enhancement in open-vocabulary tasks (2023)

\bibitem{cap-metric-lin2004rouge-l}
Lin, C.Y., Och, F.J.: Automatic evaluation of machine translation quality using longest common subsequence and skip-bigram statistics. In: Proceedings of the 42nd Annual Meeting of the Association for Computational Linguistics (ACL-04). pp. 605--612 (2004)

\bibitem{bg-transfer-aug-SIM}
Lin, J., Song, C., He, K., Wang, L., Hopcroft, J.E.: Nesterov accelerated gradient and scale invariance for adversarial attacks. arXiv preprint arXiv:1908.06281  (2019)

\bibitem{dataset-lin2014microsoftcoco}
Lin, T.Y., Maire, M., Belongie, S., Hays, J., Perona, P., Ramanan, D., Doll{\'a}r, P., Zitnick, C.L.: Microsoft coco: Common objects in context. In: Computer Vision--ECCV 2014: 13th European Conference, Zurich, Switzerland, September 6-12, 2014, Proceedings, Part V 13. pp. 740--755. Springer (2014)

\bibitem{bg-ic-vqa-liu2023llava}
Liu, H., Li, C., Wu, Q., Lee, Y.J.: Visual instruction tuning. arXiv preprint arXiv:2304.08485  (2023)

\bibitem{bg-adv-transfer-liu2016delving}
Liu, Y., Chen, X., Liu, C., Song, D.: Delving into transferable adversarial examples and black-box attacks. arXiv preprint arXiv:1611.02770  (2016)

\bibitem{liu2022convnet}
Liu, Z., Mao, H., Wu, C.Y., Feichtenhofer, C., Darrell, T., Xie, S.: A convnet for the 2020s. Proceedings of the IEEE/CVF Conference on Computer Vision and Pattern Recognition (CVPR)  (2022)

\bibitem{loshchilov2017adamw}
Loshchilov, I., Hutter, F.: Decoupled weight decay regularization. arXiv preprint arXiv:1711.05101  (2017)

\bibitem{bg-transfer-vlp-lu2023sga}
Lu, D., Wang, Z., Wang, T., Guan, W., Gao, H., Zheng, F.: Set-level guidance attack: Boosting adversarial transferability of vision-language pre-training models (2023)

\bibitem{lu2019vilbert}
Lu, J., Batra, D., Parikh, D., Lee, S.: Vilbert: Pretraining task-agnostic visiolinguistic representations for vision-and-language tasks. Advances in neural information processing systems  \textbf{32} (2019)

\bibitem{bg-transfer-crosstask-lu2020dispersion-reduction}
Lu, Y., Jia, Y., Wang, J., Li, B., Chai, W., Carin, L., Velipasalar, S.: Enhancing cross-task black-box transferability of adversarial examples with dispersion reduction. In: Proceedings of the IEEE/CVF Conference on Computer Vision and Pattern Recognition. pp. 940--949 (2020)

\bibitem{transfer-prompt-luo2023image}
Luo, H., Gu, J., Liu, F., Torr, P.: An image is worth 1000 lies: Transferability of adversarial images across prompts on vision-language models. In: The Twelfth International Conference on Learning Representations (2023)

\bibitem{bg-ovs-sparse-Luo2023SegCLIP}
Luo, H., Bao, J., Wu, Y., He, X., Li, T.: {SegCLIP}: Patch aggregation with learnable centers for open-vocabulary semantic segmentation. ICML  (2023)

\bibitem{bg-ovd-ma2023codet}
Ma, C., Jiang, Y., Wen, X., Yuan, Z., Qi, X.: Codet: Co-occurrence guided region-word alignment for open-vocabulary object detection. In: Advances in Neural Information Processing Systems (2023)

\bibitem{bg-ovd-Maaz2022Multimodal-object-centric-ovd}
Maaz, M., Rasheed, H., Khan, S., Khan, F.S., Anwer, R.M., Yang, M.H.: Class-agnostic object detection with multi-modal transformer. In: 17th European Conference on Computer Vision (ECCV). Springer (2022)

\bibitem{bg-adv-madry2017pgd}
Madry, A., Makelov, A., Schmidt, L., Tsipras, D., Vladu, A.: Towards deep learning models resistant to adversarial attacks. arXiv preprint arXiv:1706.06083  (2017)

\bibitem{bg-transfer-surrogate-malik2022adversarial}
Malik, H.S., Kunhimon, S.K., Naseer, M., Khan, S., Khan, F.S.: Adversarial pixel restoration as a pretext task for transferable perturbations  (2022)

\bibitem{bg-ic-mokady2021clipcap}
Mokady, R., Hertz, A., Bermano, A.H.: Clipcap: Clip prefix for image captioning. arXiv preprint arXiv:2111.09734  (2021)

\bibitem{dataset-mottaghi2014pascal-context}
Mottaghi, R., Chen, X., Liu, X., Cho, N.G., Lee, S.W., Fidler, S., Urtasun, R., Yuille, A.: The role of context for object detection and semantic segmentation in the wild. In: Proceedings of the IEEE conference on computer vision and pattern recognition. pp. 891--898 (2014)

\bibitem{bg-ovs-sparse-mukhoti2023open}
Mukhoti, J., Lin, T.Y., Poursaeed, O., Wang, R., Shah, A., Torr, P.H., Lim, S.N.: Open vocabulary semantic segmentation with patch aligned contrastive learning. In: Proceedings of the IEEE/CVF Conference on Computer Vision and Pattern Recognition. pp. 19413--19423 (2023)

\bibitem{bg-transfer-generative-naseer2019cross}
Naseer, M.M., Khan, S.H., Khan, M.H., Shahbaz~Khan, F., Porikli, F.: Cross-domain transferability of adversarial perturbations. Advances in Neural Information Processing Systems  \textbf{32} (2019)

\bibitem{bg-transfer-generative-naseer2021generating}
Naseer, M., Khan, S., Hayat, M., Khan, F.S., Porikli, F.: On generating transferable targeted perturbations. In: Proceedings of the IEEE/CVF International Conference on Computer Vision. pp. 7708--7717 (2021)

\bibitem{bg-transfer-crosstask-naseer2018task}
Naseer, M., Khan, S.H., Rahman, S., Porikli, F.: Task-generalizable adversarial attack based on perceptual metric. arXiv preprint arXiv:1811.09020  (2018)

\bibitem{bg-transfer-surrogate-naseer2021improving}
Naseer, M., Ranasinghe, K., Khan, S., Khan, F.S., Porikli, F.: On improving adversarial transferability of vision transformers. arXiv preprint arXiv:2106.04169  (2021)

\bibitem{bg-advdownstream-nern2024transfer}
Nern, L.F., Raj, H., Georgi, M.A., Sharma, Y.: On transfer of adversarial robustness from pretraining to downstream tasks. Advances in Neural Information Processing Systems  \textbf{36} (2024)

\bibitem{vqa-ling-shortcut-niu2021counterfactual}
Niu, Y., Tang, K., Zhang, H., Lu, Z., Hua, X.S., Wen, J.R.: Counterfactual vqa: A cause-effect look at language bias. In: Proceedings of the IEEE/CVF Conference on Computer Vision and Pattern Recognition. pp. 12700--12710 (2021)

\bibitem{bg-vlpadv-noever2021reading}
Noever, D.A., Noever, S.E.M.: Reading isn't believing: Adversarial attacks on multi-modal neurons. arXiv preprint arXiv:2103.10480  (2021)

\bibitem{bg-ic-nukrai2022CapDec}
Nukrai, D., Mokady, R., Globerson, A.: Text-only training for image captioning using noise-injected clip. arXiv preprint arXiv:2211.00575  (2022)

\bibitem{bg-adv-transfer-papernot2016transferability}
Papernot, N., McDaniel, P., Goodfellow, I.: Transferability in machine learning: from phenomena to black-box attacks using adversarial samples. arXiv preprint arXiv:1605.07277  (2016)

\bibitem{cap-metric-papineni2002bleu}
Papineni, K., Roukos, S., Ward, T., Zhu, W.J.: Bleu: a method for automatic evaluation of machine translation. In: Proceedings of the 40th annual meeting of the Association for Computational Linguistics. pp. 311--318 (2002)

\bibitem{bg-transfer-generative-poursaeed2018GAP}
Poursaeed, O., Katsman, I., Gao, B., Belongie, S.: Generative adversarial perturbations. In: Proceedings of the IEEE Conference on Computer Vision and Pattern Recognition. pp. 4422--4431 (2018)

\bibitem{bg-foundationsafety-qi2023fine}
Qi, X., Zeng, Y., Xie, T., Chen, P.Y., Jia, R., Mittal, P., Henderson, P.: Fine-tuning aligned language models compromises safety, even when users do not intend to! arXiv preprint arXiv:2310.03693  (2023)

\bibitem{bg-transfer-optim-qin2022boosting}
Qin, Z., Fan, Y., Liu, Y., Shen, L., Zhang, Y., Wang, J., Wu, B.: Boosting the transferability of adversarial attacks with reverse adversarial perturbation. arXiv preprint arXiv:2210.05968  (2022)

\bibitem{Radford2021CLIP}
Radford, A., Kim, J.W., Hallacy, C., Ramesh, A., Goh, G., Agarwal, S., Sastry, G., Askell, A., Mishkin, P., Clark, J., Krueger, G., Sutskever, I.: Learning transferable visual models from natural language supervision. In: ICML (2021)

\bibitem{bg-ic-ramos2023SmallCap}
Ramos, R., Martins, B., Elliott, D., Kementchedjhieva, Y.: Smallcap: lightweight image captioning prompted with retrieval augmentation. In: Proceedings of the IEEE/CVF Conference on Computer Vision and Pattern Recognition. pp. 2840--2849 (2023)

\bibitem{bg-ovd-Hanoona2022Bridging-object-centric-ovd}
Rasheed, H., Maaz, M., Khattak, M.U., Khan, S., Khan, F.S.: Bridging the gap between object and image-level representations for open-vocabulary detection. In: 36th Conference on Neural Information Processing Systems (NIPS) (2022)

\bibitem{dataset-ren2015coco-qa}
Ren, M., Kiros, R., Zemel, R.: Advances in neural information processing systems  \textbf{28} (2015)

\bibitem{bg-advdownstream-rezaei2019target}
Rezaei, S., Liu, X.: A target-agnostic attack on deep models: Exploiting security vulnerabilities of transfer learning. arXiv preprint arXiv:1904.04334  (2019)

\bibitem{bg-transfer-feature-salzmann2021learning}
Salzmann, M., et~al.: Learning transferable adversarial perturbations. Advances in Neural Information Processing Systems  \textbf{34},  13950--13962 (2021)

\bibitem{bg-vlpadv-schlarmann2024robust}
Schlarmann, C., Singh, N.D., Croce, F., Hein, M.: Robust clip: Unsupervised adversarial fine-tuning of vision embeddings for robust large vision-language models. arXiv preprint arXiv:2402.12336  (2024)

\bibitem{bg-transfer-vlp-shayegani2023jailbreak}
Shayegani, E., Dong, Y., Abu-Ghazaleh, N.: Jailbreak in pieces: Compositional adversarial attacks on multi-modal language models (2023)

\bibitem{bg-ic-vqa-shen2021clip-vil}
Shen, S., Li, L.H., Tan, H., Bansal, M., Rohrbach, A., Chang, K.W., Yao, Z., Keutzer, K.: How much can clip benefit vision-and-language tasks? arXiv preprint arXiv:2107.06383  (2021)

\bibitem{vqa-multi-shortcut-si2022language}
Si, Q., Meng, F., Zheng, M., Lin, Z., Liu, Y., Fu, P., Cao, Y., Wang, W., Zhou, J.: Language prior is not the only shortcut: A benchmark for shortcut learning in vqa. arXiv preprint arXiv:2210.04692  (2022)

\bibitem{bg-transfer-generative-song2018AC-GAN}
Song, Y., Shu, R., Kushman, N., Ermon, S.: Constructing unrestricted adversarial examples with generative models. Advances in Neural Information Processing Systems  \textbf{31} (2018)

\bibitem{bg-ovs-trainingfree-sun2023clip}
Sun*, S., Li*, R., Torr, P., Gu, X., Li, S.: Clip as rnn: Segment countless visual concepts without training endeavor (Dec 2023)

\bibitem{bg-adv-szegedy2013intriguing}
Szegedy, C., Zaremba, W., Sutskever, I., Bruna, J., Erhan, D., Goodfellow, I., Fergus, R.: Intriguing properties of neural networks. arXiv preprint arXiv:1312.6199  (2013)

\bibitem{tan2019lxmert}
Tan, H., Bansal, M.: Lxmert: Learning cross-modality encoder representations from transformers. arXiv preprint arXiv:1908.07490  (2019)

\bibitem{apx-semantic-attack-tao2018attacks}
Tao, G., Ma, S., Liu, Y., Zhang, X.: Attacks meet interpretability: Attribute-steered detection of adversarial samples. Advances in Neural Information Processing Systems  \textbf{31} (2018)

\bibitem{bg-foundationsafety-tu2023many}
Tu, H., Cui, C., Wang, Z., Zhou, Y., Zhao, B., Han, J., Zhou, W., Yao, H., Xie, C.: How many unicorns are in this image? a safety evaluation benchmark for vision llms. arXiv preprint arXiv:2311.16101  (2023)

\bibitem{cap-metric-vedantam2015cider}
Vedantam, R., Lawrence~Zitnick, C., Parikh, D.: Cider: Consensus-based image description evaluation. In: Proceedings of the IEEE conference on computer vision and pattern recognition. pp. 4566--4575 (2015)

\bibitem{bg-advdownstream-wang2018great}
Wang, B., Yao, Y., Viswanath, B., Zheng, H., Zhao, B.Y.: With great training comes great vulnerability: Practical attacks against transfer learning. In: 27th USENIX security symposium (USENIX Security 18). pp. 1281--1297 (2018)

\bibitem{bg-transfer-optim-wang2021enhancing}
Wang, X., He, K.: Enhancing the transferability of adversarial attacks through variance tuning. In: Proceedings of the IEEE/CVF Conference on Computer Vision and Pattern Recognition. pp. 1924--1933 (2021)

\bibitem{bg-transfer-feature-wang2021feature}
Wang, Z., Guo, H., Zhang, Z., Liu, W., Qin, Z., Ren, K.: Feature importance-aware transferable adversarial attacks. In: Proceedings of the IEEE/CVF international conference on computer vision. pp. 7639--7648 (2021)

\bibitem{methods-motivation-waseda2023closer}
Waseda, F., Nishikawa, S., Le, T.N., Nguyen, H.H., Echizen, I.: Closer look at the transferability of adversarial examples: How they fool different models differently. In: Proceedings of the IEEE/CVF Winter Conference on Applications of Computer Vision. pp. 1360--1368 (2023)

\bibitem{bg-ovd-wu2023cora}
Wu, X., Zhu, F., Zhao, R., Li, H.: Cora: Adapting clip for open-vocabulary detection with region prompting and anchor pre-matching. ArXiv  \textbf{abs/2303.13076} (2023)

\bibitem{bg-ovs-dense-xie2023sed}
Xie, B., Cao, J., Xie, J., Khan, F.S., Pang, Y.: Sed: A simple encoder-decoder for open-vocabulary semantic segmentation (2023)

\bibitem{bg-transfer-aug-DIM}
Xie, C., Zhang, Z., Zhou, Y., Bai, S., Wang, J., Ren, Z., Yuille, A.L.: Improving transferability of adversarial examples with input diversity. In: Proceedings of the IEEE/CVF Conference on Computer Vision and Pattern Recognition. pp. 2730--2739 (2019)

\bibitem{bg-transfer-optim-xiong2022stochastic}
Xiong, Y., Lin, J., Zhang, M., Hopcroft, J.E., He, K.: Stochastic variance reduced ensemble adversarial attack for boosting the adversarial transferability. In: Proceedings of the IEEE/CVF Conference on Computer Vision and Pattern Recognition. pp. 14983--14992 (2022)

\bibitem{bg-ovs-dense-xu2023san}
Xu, M., Zhang, Z., Wei, F., Hu, H., Bai, X.: San: Side adapter network for open-vocabulary semantic segmentation. IEEE Transactions on Pattern Analysis and Machine Intelligence  (2023)

\bibitem{bg-ovs-dense-xu2022zssegbaseline}
Xu, M., Zhang, Z., Wei, F., Lin, Y., Cao, Y., Hu, H., Bai, X.: A simple baseline for open-vocabulary semantic segmentation with pre-trained vision-language model. In: European Conference on Computer Vision. pp. 736--753. Springer (2022)

\bibitem{bg-transfer-optim-xu2022adversarially}
Xu, X., Zhang, J.Y., Ma, E., Son, H.H., Koyejo, S., Li, B.: Adversarially robust models may not transfer better: Sufficient conditions for domain transferability from the view of regularization. In: International Conference on Machine Learning. pp. 24770--24802. PMLR (2022)

\bibitem{bg-ovs-dense-xu2023masqclip}
Xu, X., Xiong, T., Ding, Z., Tu, Z.: Masqclip for open-vocabulary universal image segmentation. In: Proceedings of the IEEE/CVF International Conference on Computer Vision. pp. 887--898 (2023)

\bibitem{bg-advdownstream-yamada2022does}
Yamada, Y., Otani, M.: Does robustness on imagenet transfer to downstream tasks? In: Proceedings of the IEEE/CVF Conference on Computer Vision and Pattern Recognition. pp. 9215--9224 (2022)

\bibitem{bg-transfer-generative-yang2022C-GSP}
Yang, X., Dong, Y., Pang, T., Su, H., Zhu, J.: Boosting transferability of targeted adversarial examples via hierarchical generative networks. In: European Conference on Computer Vision. pp. 725--742. Springer (2022)

\bibitem{bg-ovs-sparse-yi2023simseg}
Yi, M., Cui, Q., Wu, H., Yang, C., Yoshie, O., Lu, H.: A simple framework for text-supervised semantic segmentation. In: Proceedings of the IEEE/CVF Conference on Computer Vision and Pattern Recognition (CVPR). pp. 7071--7080 (2023)

\bibitem{bg-ovs-dense-yu2023fcclip}
Yu, Q., He, J., Deng, X., Shen, X., Chen, L.C.: Convolutions die hard: Open-vocabulary segmentation with single frozen convolutional clip. In: NeurIPS (2023)

\bibitem{zang2022OV-DETR}
Zang, Y., Li, W., Zhou, K., Huang, C., Loy, C.C.: Open-vocabulary detr with conditional matching (2022)

\bibitem{bg-vlpadv-zhang2022towards}
Zhang, J., Yi, Q., Sang, J.: Towards adversarial attack on vision-language pre-training models. In: Proceedings of the 30th ACM International Conference on Multimedia. pp. 5005--5013 (2022)

\bibitem{bg-transfer-optim-zhang2023vit-token-gradient-regularizatione}
Zhang, J., Huang, Y., Wu, W., Lyu, M.R.: Transferable adversarial attacks on vision transformers with token gradient regularization. In: Proceedings of the IEEE/CVF Conference on Computer Vision and Pattern Recognition. pp. 16415--16424 (2023)

\bibitem{bg-advdownstream-zhang2022remos}
Zhang, Z., Li, Y., Wang, J., Liu, B., Li, D., Guo, Y., Chen, X., Liu, Y.: Remos: reducing defect inheritance in transfer learning via relevant model slicing. In: Proceedings of the 44th International Conference on Software Engineering. pp. 1856--1868 (2022)

\bibitem{bg-transfer-surrogate-zhao2023minimizing}
Zhao, A., Chu, T., Liu, Y., Li, W., Li, J., Duan, L.: Minimizing maximum model discrepancy for transferable black-box targeted attacks. In: Proceedings of the IEEE/CVF Conference on Computer Vision and Pattern Recognition. pp. 8153--8162 (2023)

\bibitem{bg-ovd-zhao2022exploiting-VL-PLM}
Zhao, S., Zhang, Z., Schulter, S., Zhao, L., Vijay~Kumar, B., Stathopoulos, A., Chandraker, M., Metaxas, D.N.: Exploiting unlabeled data with vision and language models for object detection. In: ECCV. pp. 159--175. Springer (2022)

\bibitem{bg-transfer-vlp-zhao2024evaluating}
Zhao, Y., Pang, T., Du, C., Yang, X., Li, C., Cheung, N.M.M., Lin, M.: On evaluating adversarial robustness of large vision-language models. Advances in Neural Information Processing Systems  \textbf{36} (2024)

\bibitem{bg-transfer-zhao2023revisiting}
Zhao, Z., Zhang, H., Li, R., Sicre, R., Amsaleg, L., Backes, M., Li, Q., Shen, C.: Revisiting transferable adversarial image examples: Attack categorization, evaluation guidelines, and new insights. arXiv preprint arXiv:2310.11850  (2023)

\bibitem{bg-ovd-zhou2022detic}
Zhou, X., Girdhar, R., Joulin, A., Kr{\"a}henb{\"u}hl, P., Misra, I.: Detecting twenty-thousand classes using image-level supervision. In: ECCV (2022)

\bibitem{bg-vlpadv-zhou2023advclip}
Zhou, Z., Hu, S., Li, M., Zhang, H., Zhang, Y., Jin, H.: Advclip: Downstream-agnostic adversarial examples in multimodal contrastive learning. In: Proceedings of the 31st ACM International Conference on Multimedia. pp. 6311--6320 (2023)

\bibitem{bg-transfer-vlp-zou2023universal}
Zou, A., Wang, Z., Carlini, N., Nasr, M., Kolter, J.Z., Fredrikson, M.: Universal and transferable adversarial attacks on aligned language models (2023)

\bibitem{bg-vlpadv-zou2023universal}
Zou, A., Wang, Z., Kolter, J.Z., Fredrikson, M.: Universal and transferable adversarial attacks on aligned language models. arXiv preprint arXiv:2307.15043  (2023)

\bibitem{bg-transfer-aug-RDI}
Zou, J., Pan, Z., Qiu, J., Liu, X., Rui, T., Li, W.: Improving the transferability of adversarial examples with resized-diverse-inputs, diversity-ensemble and region fitting. In: Computer Vision--ECCV 2020: 16th European Conference, Glasgow, UK, August 23--28, 2020, Proceedings, Part XXII. pp. 563--579. Springer (2020)

\end{thebibliography}

\newpage
\appendix

{ \bfseries\boldmath
\centering \Large{Supplementary Materials}\\
}

% \flushleft
\section{Pseudocode}\label{apx:algorithm}
We use $\mathcal{F}_l(x), \mathcal{F}_l(x')$ to denote intermediate encoder features obtained from layer $l$. Each $\mathcal{F}$  can be expanded as a set of representations of individual image patches $\mathcal{F}_l(x) = \{f^0_l, f^1_l, \dots, f^{\lceil\frac{HW}{d^2}\rceil}_l\} = \{f^p_l \in \mathbb{R}^{N} | p\in[0,\lceil\frac{HW}{d^2}\rceil]\}$, which are traversed through by the inner for loop.
\begin{algorithm}[]
\small
\caption{Patch Representation Misalignment (PRM) Attack}\label{alg:prm}
\begin{algorithmic}[1]
\Require Surrogate vision encoder $\mathcal{F}$ with attackable layers $L$, perturbation budget $\epsilon$, clean image $x$, learning rate $\alpha$ and number of iterations $N$ %, range of scaling factors $(a, b)$
% \Ensure $y = x^n$
\State $\delta_0 \sim \mathcal{U}(-\epsilon, \epsilon)$; $x'_0 \gets$ Clip($x+\delta_0, 0, 255$) \Comment{Initialise perturbation and adversary}
\For{$t \gets 0$ to $N$}
    % \State scale $\sim \mathcal{U}(a, b)$ \Comment{Randomly resize clean and adversarial samples}
    % \State $x_t \gets$ Resize($x$, scale); $x'_t \gets$ Resize($x'_t$, scale) 
    % \Comment{Randomly rescale inputs}
    \State $\mathcal{L}_\mathrm{PRM} = 0$ 
    \For{$l \in L$}
    \Comment{Traverse through all layers}
    % \State $\{f^p_{l}\}_{p=0}^{\lceil HW/d^2\rceil} = \mathcal{F}_l(x_t)$
    % \Comment{Collect clean patch embeddings from layer $l$}
    % \State $\{f'^p_{l}\}_{p=0}^{\lceil HW/d^2\rceil} = \mathcal{F}_l(x_t')$  \Comment{Collect adversarial patch embeddings from layer $l$}
    % \State $\mathcal{L}_\mathrm{PRM}$ += $\sum_{p=0}^{\lceil HW/d^2\rceil } ({f}^p_l \cdot {f'}^p_l) / (\|{f}^p_l \| \|{f'}^p_l\|) $  \Comment{Compute loss for all tokens}
    \For{$(f, f')$ \textbf{in} 
    $(\mathcal{F}_l(x), \mathcal{F}_l(x_t'))$}
    \Comment{Traverse through all patches}
    \State $\mathcal{L}_\mathrm{PRM}$ += $(f \cdot f')/(\|f\|\|f'\|)$ 
    \EndFor \Comment{Accumulate loss for all patch representations in layer $l$}
    \EndFor \Comment{Accumulate loss for all attackable encoder layers}
    \State $\delta_{t+1} \gets \delta_{t} - \alpha \cdot \mathrm{Sign} (\nabla_{\delta_{t}} \mathcal{L}_\mathrm{PRM})$ \Comment{Update adversarial perturbation}
    \State $\delta_{t+1} \gets$ Clip($\delta_{t+1}, -\epsilon, \epsilon$) \Comment{Project perturbation to $\epsilon$-ball}
    \State $x'_{t+1} \gets$ Clip($x+\delta_{t+1}, 0, 255$) \Comment{Project adversary to valid image range}
\EndFor
\end{algorithmic}
\end{algorithm}
\vspace{-30pt}
\section{Ablation Studies: Results and Further Discussions}
In this section, we present the quantitative results of abaltaion studies mentioned in Sec. \ref{subsec:ablation-cosine-dim}. All evaluations are done on COCO-Stuff validation set. We present the performance metrics of a subset of OVS models though the trend in performance generalises to other tasks and datasets. 

\subsection{Cosine Similarity-Based Distortion}\label{apx:subsec:flatten}

We hypothesized, in Sec. \ref{subsec:prelim}, that angular distance could be a favourable distance metric in the intermediate feature space of visino-language alignment pre-trained encoders.
To check this assumption, we repeat our experiments using architecturally identical non-CLIP vision encoders trained with classification objectives as surrogates. As shown in Table \ref{tab:cosine-nonclip}, while PRM still outperforms baselines, their performance gap shrinks notably (e.g. on selected OVS victim models, the performance gap between PRM and MSE-based NRD shrinks from 23\% on CLIP vision encoders to around 11\% on classification-trained encoders). This suggests that vision-language alignment pre-training could have played a role in assigning greater semantic importance to directional information over magnitude or variance (in comparison to classification cross entropy-trained alternatives).

\begin{table}[!h]
\centering
\fontsize{6}{8}\selectfont
    \caption{Does vision-language alignment pre-training contribute to the emphasis on directional information in the feature space? The fact that cosine-similarity-based PRM has a larger performance gain over NRD (an MSE-based alternative) on alignment-pretrained encoders suggests this is likely a valid assumption.}
    \label{tab:cosine-nonclip}
    \begin{tabular}{cc|C{1cm}|C{1cm}|C{1.1cm}|C{1.1cm}|C{1.1cm}|C{1.1cm}|C{1.1cm}}
    \specialrule{.8pt}{1pt}{1pt}
     &  & \multicolumn{2}{c|}{CAT-Seg}  & \multicolumn{2}{c|}{SED}  & \multicolumn{3}{c}{FC-CLIP} \\
     \cline{3-9}
     & \multirow{-2}{*}{$\mathcal{L}_{attack}$} & ViT-B & ViT-L & CNeXt-B & CNeXt-L & RN50 & RN50$\times$64 & CNeXt-L \\
      \cline{2-9}
    \multirow{-3}{*}{\backslashbox[8mm]{S}{T}} 
    & Clean & 46.17 & 50.35 & 46.21 & 49.57 & 54.85 & 60.53 & 63.22 \\
     \hline
     & NRD & 27.25 & 44.23 & 34.88 & 42.53 & 39.43 & 52.46 & 55.52\\
     & \textbf{PRM} & 4.23 & 22.44 & 11.57 & 22.49 & 13.25 & 30.90 & 34.39\\
    \multirow{-3}{*}{\begin{tabular}[c]{@{}c@{}}CLIP\\ ViT-B\end{tabular}} & $\mathrm{\Delta}$ &23.02 & 21.79 & 23.31 & 20.04 & 26.18 & 21.55 & 21.13\\
    \hline 
     & NRD & 35.39 & 43.45 & 36.24 & 42.07 & 43.13 & 53.61 & 56.83\\
     & \textbf{PRM} & 24.44 & 36.92 & 25.47 & 35.46 & 33.57 & 49.56 & 52.37\\
    \multirow{-3}{*}{\begin{tabular}[c]{@{}c@{}}Classification\\ ViT-B\end{tabular}} & $\mathrm{\Delta}$ & 10.96 & 6.53 & 10.77 & 6.60 & 9.56 & 4.05 & 4.46\\
    \specialrule{.8pt}{1pt}{1pt}
    \end{tabular}
\end{table}

To further validate the efficacy of our loss design, we individually examine the two core components of our loss: (1) the patch-wise approach and (2) the cosine similarity-based distortion. As shown in Table \ref{tab:ablation-flatten}, the combination of them provides a further performance improvement.
\begin{table}[!h]
\centering
\fontsize{6}{8}\selectfont
    \caption{To flatten or not? Through this set of ablation studies, we verified that (1) the patch-wise approach tends to be more effective than the global flattened approach when both are similarly based on cosine similarity; (2) cosine-similarity-based distortion (PRM) is more effective than MSE-based alternative (NRD)  when both use a similar global (non patch-wise) approach.}
    \label{tab:ablation-flatten}
    \begin{tabular}{cc|C{1cm}|C{1cm}|C{1.1cm}|C{1.1cm}|C{1.1cm}|C{1.1cm}|C{1.1cm}}
    \specialrule{.8pt}{1pt}{1pt}
     &  & \multicolumn{2}{c|}{CAT-Seg}  & \multicolumn{2}{c|}{SED}  & \multicolumn{3}{c}{FC-CLIP} \\
     \cline{3-9}
     & \multirow{-2}{*}{$\mathcal{L}_{attack}$} & ViT-B & ViT-L & CNeXt-B & CNeXt-L & RN50 & RN50$\times$64 & CNeXt-L \\
      \cline{2-9}
    \multirow{-3}{*}{\backslashbox[8mm]{S}{T}} 
    & Clean & 46.17 & 50.35 & 46.21 & 49.57 & 54.85 & 60.53 & 63.22 \\
     \hline \hline
     & \textbf{Patch-PRM} & \bluebf{4.23} & \bluebf{22.44} & \bluebf{11.57} & \bluebf{22.49} & \bluebf{13.25} & \bluebf{30.90} & \bluebf{34.39}\\
      & Global-PRM & 12.07 & 34.64 & 21.27 & 32.90 & 22.63 & 41.82 & 45.38\\
    \multirow{-3}{*}{\begin{tabular}[c]{@{}c@{}}CLIP\\ ViT-B\end{tabular}}
    & Global-NRD & 27.25 & 44.23 & 34.88 & 42.53 & 39.43 & 52.46 & 55.52\\
    \hline \hline
     & \textbf{Patch-PRM} & \bluebf{23.34} & \bluebf{32.09} & \bluebf{4.35} & \bluebf{5.65} & \bluebf{5.30} & \bluebf{7.07} & \bluebf{0.62}\\
     & Global-PRM & 24.63 & 33.29 & 5.22 & 7.00 & 6.70 & 8.72 & 0.66\\
    \multirow{-3}{*}{\begin{tabular}[c]{@{}c@{}}CLIP\\ {CNeXt-L}\end{tabular}} 
    & Global-NRD & 39.06 & 45.17 & 27.41 & 29.51 & 32.06 & 37.78 & 19.60\\
    \specialrule{.8pt}{1pt}{1pt}
    \end{tabular}
\end{table}
In particular, we attempt an alternative usage of cosine similarity that flattens features before cosine similarity computation (i.e. globally distorting the entire volumetric feature using cosine similarity). While this configuration still outperforms MSE-based and variance-based baselines, it is not as performant as our default patch-wise setup which treats every patch descriptor as an independent sample and computes cosine similarity specifically along the embedding (ViT) or channel (CNN) dimensions of the features. 
This drop in performance indicates that the patch-wise approach tends to be more effective than the global flattened approach even when both are similarly based on cosine similarity.
Moreover, the fact that Global-PRM still outperforms MSE-based alternative (NRD) provide further evidence that cosine-similarity based approach contributes to stronger attack performance.

\subsection{Which features to attack}\label{apx:subsec:whichfeatures}
As mentioned in Sec. \ref{subsec:setup}, we choose to attack all tokens (or patches) from all vision encoder layers. This contrasts with previous studies that argue single-layer attacks (only distorting features from one surrogate layer) on mid-layer features would be sufficient \cite{bg-transfer-crosstask-naseer2018task,bg-transfer-crosstask-lu2020dispersion-reduction,bg-transfer-feature-salzmann2021learning}. To determine whether attacking all features is necessary, we attempt several alternatives that attack subsets of features and present results in Table \ref{tab:ablation-whichfeatures} below. 
\begin{enumerate}
    \item Only attacking \texttt{CLS} tokens from all layers (ViT-B surrogate). 
    \item Only attacking the final layer features (ViT-B/ConvNeXt-L surrogates). 
    \item Only attacking mid-layer features (Layer 6 in ViT-B surrogates; Layer 6 of the first ConvNeXt Stage in ConvNeXt-L surrogates). 
\end{enumerate}\vspace{-20pt}
\begin{table}[!h]
\centering
\fontsize{6}{8}\selectfont
\caption{Which features to Attack? All experiments uses $\mathcal{L}_\mathrm{PRM}$}  \label{tab:ablation-whichfeatures}
\begin{tabular}{cc|C{1cm}|C{1cm}|C{1.1cm}|C{1.1cm}|C{1.1cm}|C{1.1cm}|C{1.1cm}}
\specialrule{.8pt}{1pt}{1pt}
 &  & \multicolumn{2}{c|}{CAT-Seg}  & \multicolumn{2}{c|}{SED}  & \multicolumn{3}{c}{FC-CLIP} \\
 \cline{3-9}
 & \multirow{-2}{*}{Layer/Token} & ViT-B & ViT-L & CNeXt-B & CNeXt-L & RN50 & RN50$\times$64 & CNeXt-L \\
  \cline{2-9}
\multirow{-3}{*}{\backslashbox[8mm]{S}{T}} 
& Clean & 46.17 & 50.35 & 46.21 & 49.57 & 54.85 & 60.53 & 63.22 \\
 \hline\hline
 & \textbf{AllLayers} & \bluebf{4.23} & \bluebf{22.44} & \bluebf{11.57} & \bluebf{22.49} & \bluebf{13.25} & \bluebf{30.90} & \bluebf{34.39}\\
 & \texttt{CLS}-only & 22.24 & 41.42 & 29.98 & 39.54 & 34.29 & 49.36 & 54.08\\
 & MidLayer & 10.19 & 30.77 & 15.43 & 28.62 &18.42 & 37.34 & 41.62\\
\multirow{-4}{*}{\begin{tabular}[c]{@{}c@{}}CLIP\\ ViT-B\end{tabular}} & LastLayer & 11.18 & 34.97 & 23.76 & 34.66 & 26.33 & 43.72 & 48.26\\
 \hline \hline
 & \textbf{AllLayers} & \bluebf{23.34} & \bluebf{32.09} & \bluebf{4.35} & \bluebf{5.65} & \bluebf{5.30} & \bluebf{7.07} & \bluebf{0.62} \\
 & MidLayer & 38.17 & 46.00 & 26.25 & 38.26 &  25.07 & 40.80 & 47.49\\
\multirow{-3}{*}{\begin{tabular}[c]{@{}c@{}}CLIP\\ {CNeXt-L}\end{tabular}} & LastLayer & 34.42 & 41.23 & 16.94 & 17.72 & 18.63 & 20.15 & 2.75\\
\specialrule{.8pt}{1pt}{1pt}
\end{tabular}
\end{table}
While attacking the full feature hierarchy results in the strongest attack, we acknowledge that the observation regarding the strong generalizability of mid-layer features remains valid to some extent: mid-layer features have higher generalizability than final-layer features in ViTs. Curiously, the second-best option is the last-layer attack for CNN-based surrogates whereas the mid-layer attack is the second-best option for ViT-based surrogates. 
Based on observations of the target models, we hypothesise that the improved transferability observed in multi-layer and full-spatial coverage attacks may be attributed to the following reasons: 

\begin{enumerate}
    \item Only attacking the global \texttt{CLS} token is insufficient since many downstream models (such as \cite{bg-ovs-dense-cho2023catseg} that use CLIP dense features for cost volume computation) tend to prioritise dense patch representations over the global class token. This tendency is particularly notable in tasks involving dense predictions, such as segmentation and object detection. 
    \item The assumption that distortions in mid-layer features can propagate down the feature hierarchy to affect model decisions \cite{bg-transfer-crosstask-lu2020dispersion-reduction} may be less applicable due to more intricate modes of interactions between downstream models and pre-trained features. 
    The interactions between the CLIP encoder backbone and other components of the downstream model take many intricate forms, some downstream models go as far as using CLIP representations as prompt prefixes to query other foundation models \cite{bg-ic-li2023decap,bg-ic-nukrai2022CapDec,bg-ic-mokady2021clipcap,bg-ic-ramos2023SmallCap,bg-ic-fei-iccv23-ViECap,lu2019vilbert}. This means that we cannot assume the preservation of feature locality or spatial continuity of a model's internal representations throughout the downstream model's pipeline, which means certain feature distortions may not be propagated to the decision layer. 
    \item Some downstream models (such as \cite{bg-ovs-dense-xu2023san}) selectively use certain encoder layers for specific tasks. Consequently, in a downstream model, inputs may not undergo the full forward pass of a pre-trained vision encoder; instead, only specific layers of the vision encoder's intermediate features are extracted for further processing. This implies that attackers cannot assume which intermediate features are important.
    \item The hierarchy of multiple latent spaces from various layers of a vision encoder can be interpreted as a self-ensemble \cite{bg-transfer-surrogate-naseer2021improving}, which is believed to improve transferability by making the feature and loss landscapes more generalisable.
\end{enumerate}
Therefore, we recommend attacking all tokens from all layers alike to ensure maximum coverage of perturbations, and to increase the likelihood of impacting the features used by the victim model. 

\subsection{The effect of input scale diversity}\label{apx:subsec:scale}
In the main experiments, we employ random rescaling augmentation for all competing loss objectives. We ablate this component and show results without input scale augmentation. Our method still outperforms baselins when all are examined without such augmentation. However, while augmentation is usually considered orthogonal to loss objective design, we do notice that input diversity magnifies the performance gap between our method and the baselines.
\begin{table}[!h]
    \centering
    \fontsize{6}{8}\selectfont
        
    \caption{Attack performance on COCO-Stuff without random rescaling augmentation.}
    \label{tab:ablation-scale}
    
    \begin{tabular}{cc|C{1cm}|C{1cm}|C{1.1cm}|C{1.1cm}|C{1.1cm}|C{1.1cm}|C{1.1cm}}
    \specialrule{.8pt}{1pt}{1pt}
     &  & \multicolumn{2}{c|}{CAT-Seg}  & \multicolumn{2}{c|}{SED}  & \multicolumn{3}{c}{FC-CLIP} \\
     \cline{3-9}
     & \multirow{-2}{*}{$\mathcal{L}_{attack}$} & ViT-B & ViT-L & CNeXt-B & CNeXt-L & RN50 & RN50$\times$64 & CNeXt-L \\
      \cline{2-9}
    \multirow{-3}{*}{\backslashbox[8mm]{S}{T}} 
    & Clean & 46.17 & 50.35 & 46.21 & 49.57 & 54.85 & 60.53 & 63.22 \\
     \hline\hline
      & DR & 44.07 & 49.20 & 43.72 & 48.05 & 49.33 & 57.85 & 60.73\\
     & NRD & 43.76 & 49.21 & 43.47 & 47.97 & 48.57 & 57.84 & 60.48 \\
    \multirow{-3}{*}{\begin{tabular}[c]{@{}c@{}}CLIP\\ ViT-B\end{tabular}} & \textbf{PRM} & \bluebf{11.18} & \bluebf{34.97} & \bluebf{23.76} & \bluebf{34.66} & \bluebf{26.33} & \bluebf{43.72} & \bluebf{39.04}\\
     \hline\hline
     & DR & 44.38 & 49.47 & 42.61 & 46.43 & 45.82 & 55.58 & 4.32\\
     & NRD & 43.98 & 49.47 & 41.68 & 46.03 & 44.60 & 45.66 & 2.99\\
    \multirow{-3}{*}{\begin{tabular}[c]{@{}c@{}}CLIP\\ {CNeXt-L}\end{tabular}} & \textbf{PRM} & \bluebf{40.78} & \bluebf{47.67} & \bluebf{28.70} & \bluebf{35.35} & \bluebf{27.35} & \bluebf{41.82} & \bluebf{0.00}\\
    \specialrule{.8pt}{1pt}{1pt}
    \end{tabular}
\end{table}
This technique \cite{bg-transfer-aug-DIM,bg-transfer-aug-RDI} significantly amplifies attack efficacy for all attacks by fostering scale invariance, which makes them more adept at fooling downstream models that operate at different input resolutions. 
Indeed, among CLIP's downstream models, some require that the input image be resized to a fixed scale \cite{bg-ic-vqa-awadalla2023openflamingo} while others can accommodate dynamic scale inputs. Moreover, downstream models that handle dynamic scales but use the ViT-based CLIP vision encoder backbone employ diverse techniques (such as sliding windows \cite{bg-ovs-dense-cho2023catseg} and learnable positional embeddings \cite{bg-ovs-dense-Han2023DeOP}) to address the discrepancy between the original ViT input resolution of the pre-trained backbone and the dynamic test-time resolution. By finding perturbations that effectively disturb model behaviours at various scales, we can expose the alarming efficacy of CLIP feature-based attacks on its downstream models.

\subsection{The effect of alignment pretraining.} \label{apx:subsec:alignment-pretraining}
The fact that attacks crafted with either ViT or ConvNeXt surrogate can transfer to downstream models regardless of the target model's vision encoder architecture (e.g. that use a different vision encoder backbone, as discussed in Sec. \ref{subsec:main-results-discussions}) prompts us to consider how certain factors in the pre-training process may contribute to shared adversarial vulnerabilities across CLIP's downstream models. 

Various factors in the pre-training process (e.g. data, training objective and optimisation techniques) can contribute to the generalisability of an encoder's feature space. To examine how different pretraining protocols affect an encoder's efficacy as an attack surrogate, we perform our attack using architecturally identical vision encoders pre-trained and fine-tuned with different protocols in Table \ref{tab:ablation-alignment-pretraining}. Notably, adversaries crafted with classification-trained encoders exhibit much lower transferability compared to those crafted with a vision-language alignment-pretrained vision encoder of the same architecture, suggesting that the latter encoder serves as a better tool for identifying common non-robust features on which downstream models rely. 
We hypothesise that the combination of web-scale vision-language data, the alignment training objective and various training configurations make alignment-trained encoders better feature extractors and hence the basis of more transferable attacks.

\begin{table}[!h]
\centering
\fontsize{6}{8}\selectfont
    \caption{The effect of alignment pretraining. All experiments uses $\mathcal{L}_\mathrm{PRM}$. Adversaries crafted with classification-trained encoders exhibit much lower transferability compared to those crafted with a vision-language alignment-pretrained vision encoder of the same architecture. Further fine-tuning on ImageNet can make alignment-pre-trained off-the-shelf CLIP vision encoders even stronger attack surrogates.}
    \label{tab:ablation-alignment-pretraining}
    \begin{tabular}{cc|C{1cm}|C{1cm}|C{1.1cm}|C{1.1cm}|C{1.1cm}|C{1.1cm}|C{1.1cm}}
    \specialrule{.8pt}{1pt}{1pt}
    & & \multicolumn{2}{c|}{CAT-Seg}  & \multicolumn{2}{c|}{SED}  & \multicolumn{3}{c}{FC-CLIP} \\
     \cline{3-9}
    & & ViT-B & ViT-L & CNeXt-B & CNeXt-L & RN50 & RN50$\times$64 & CNeXt-L \\
    \cline{2-9}
    \multirow{-3}{*}{\backslashbox[25mm]{S}{T}} 
     & Clean & 46.17 & 50.35 & 46.21 & 49.57 & 54.85 & 60.53 & 63.22 \\
     \hline\hline
     \multicolumn{2}{c|}{LAION V-L Align. ViT-B} & \bluebf{4.23} & \bluebf{22.44} & \bluebf{11.57} & \bluebf{22.49} & \bluebf{13.25} & \bluebf{30.90} & \bluebf{34.39}\\
     \multicolumn{2}{c|}{ImageNet Class. ViT-B }& 24.44 & 36.92 & 25.47 & 35.46 &  33.57 & 49.56 & 52.37 \\
     \hline \hline
   \multicolumn{2}{c|}{LAION V-L Align. CNeXt-L} & 23.34 & 32.09 & 4.35 & 5.65 & 5.30 & 7.07 & 0.62\\
     \multicolumn{2}{c|}{ImageNet Class. CNeXt-L }& 35.71 & 42.83 & 19.37 & 29.32 & 18.06 & 31.93 & 40.40 \\
     \multicolumn{2}{c|}{LAION+ImageNetFT CNeXt-L } & \bluebf{21.32} & \bluebf{28.72} & \bluebf{3.82} & \bluebf{5.14} & \bluebf{6.43} & \bluebf{8.52} & \bluebf{1.56}\\
    \specialrule{.8pt}{1pt}{1pt}
    \end{tabular}
\end{table}
\section{Additional Qualitative Results}\label{apx:qualitative}

\begin{figure}[H]
\begin{tikzpicture}[]
    \fill [fill=brightgreen,anchor=north west,inner sep=0] (2.4,6.6) rectangle (-0.05,3.15);
    \fill [fill=brightred,anchor=north west,inner sep=0] (4.8,6.6) rectangle (2.38,3.15);
    % \fill [fill=brightred!30,anchor=north west,inner sep=0] (12.1,3.3) rectangle (4.8,-3.9);
    % \fill [fill=white,anchor=north west,inner sep=0] (-0.1,3) rectangle (12.1,3.2);
    % \fill [fill=white,anchor=north west,inner sep=0] (-0.1,-.4) rectangle (12.1,-.2);
    
    % \node [anchor=north west,rotate=0]at (1,4.65) {\tiny SAN};
    % \node [anchor=north west,rotate=0]at (3.7,4.65) {\tiny DeOP};
    % \node [anchor=north west,rotate=0]at (6.5,4.65) {\tiny FC-CLIP};
    % \node [anchor=north west,rotate=0]at (9.2,4.65) {\tiny CAT-Seg-L};
    \node[anchor=north west,inner sep=0] at (0,6.5){
    \includegraphics[width=0.19\linewidth,trim={0 0 0 0cm},clip]{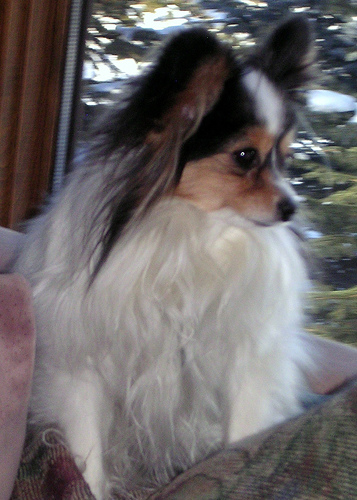}
    \includegraphics[width=0.19\linewidth,trim={0 0 0 0cm},clip]{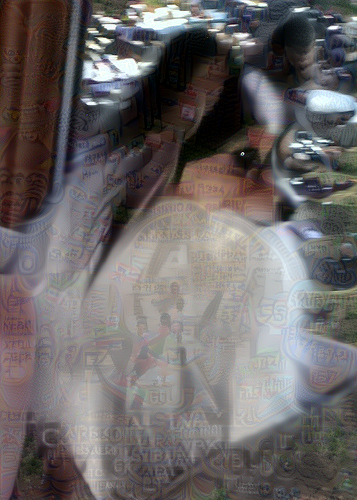}
    };
    
    \node[anchor=north west,inner sep=0] at (-0.05,2.85) {
    \includegraphics[width=\textwidth]{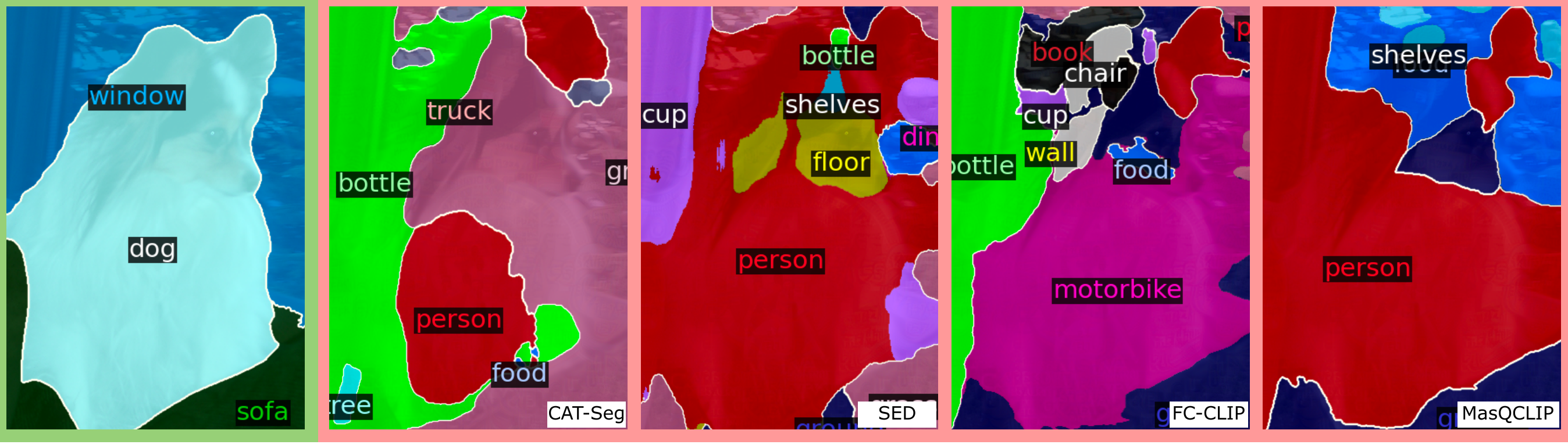}
    }; 
    
    \node[anchor=north west,inner sep=0] at (-0.05,-1.0) {
    \includegraphics[width=\textwidth]{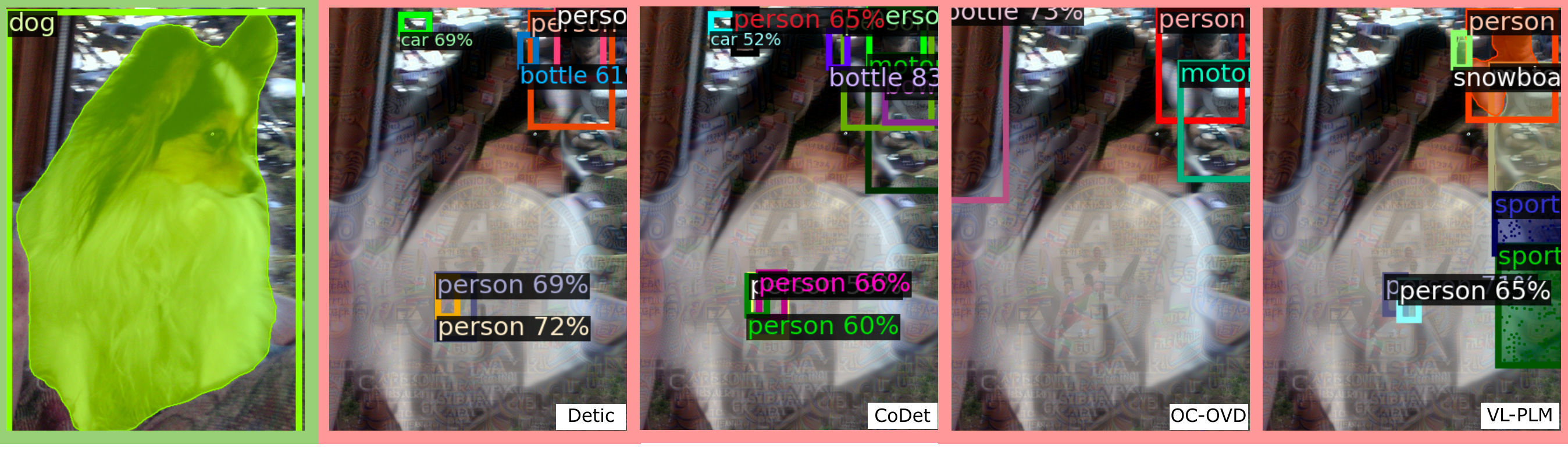}
    }; 
    
    % \node [anchor=north west,rotate=0]at (1,.3) {\tiny Detic};
    % \node [anchor=north west,rotate=0]at (3.5,.3) {\tiny OC-OVD};
    % \node [anchor=north west,rotate=0]at (6.7,.3) {\tiny CoDet};
    % \node [anchor=north west,rotate=0]at (9.2,.3) {\tiny VL-PLM};

    % \node [anchor=north west,rotate=0] at (0,-4.15) [fill, rounded corners=5pt, fill=bubblegrey!20,] {\tiny Please describe this image using one sentence.};
    \node [anchor=north west,rotate=0,align=left, text width=0.57\textwidth] at (4.9,6.5) [fill, rounded corners=5pt, fill=brightred!30,] {\tiny CapDec: \texttt{A close up of a baseball player wearing a catchers mitt.}};
    \node [anchor=north west,rotate=0,align=left, text width=0.57\textwidth] at (4.9,6.2)[fill, rounded corners=5pt, fill=brightred!30,] {\tiny LLaVa: \texttt{A person is holding a sign that says "A".}};
    \node [anchor=north west,rotate=0,align=left, text width=0.57\textwidth] at (4.9,5.9)[fill, rounded corners=5pt, fill=brightred!30,] {\tiny OpenFlamingo: \texttt{An image of the seal of the United States of America. }};
    \node [anchor=north west,rotate=0,align=left, text width=0.57\textwidth] at (4.9,5.6) [fill, rounded corners=5pt, fill=brightgreen!40,] {\tiny This image features a brown and white dog sitting on a couch.};

    \node [anchor=north west,rotate=0,align=left, text width=0.57\textwidth] at (4.9,4.8) [fill, rounded corners=5pt, fill=gray!20] {\tiny Where can you find this animal?};
    \node [anchor=north west,rotate=0,align=left, text width=0.57\textwidth] at (4.9,4.5) [fill, rounded corners=5pt, fill=brightred!30,] {\baselineskip=5pt \tiny OpenFlamingo:  \texttt{It's on the side of a building in the middle of the city.} \par};
    \node [anchor=north west,rotate=0,align=left, text width=0.57\textwidth] at (4.9,4.2) [fill, rounded corners=5pt, fill=brightred!30,]  {\baselineskip=5pt \tiny LLaVa: \texttt{The animal in the image is a giraffe. Giraffes are native to Africa and typically inhabit savannas, grasslands, and open woodlands\dots} \par};
    \node [anchor=north west,rotate=0,align=left, text width=0.57\textwidth] at (4.9,3.7) [fill, rounded corners=5pt, fill=brightgreen!40,] {\baselineskip=5pt \tiny The dog is sitting on a couch, which is likely located indoors\dots\par};

    \node [anchor=north west,rotate=0]at (4,3.2) {\scriptsize Open-Vocabulary Segmentation};
    \node [anchor=north west,rotate=0]at (4.3,-0.6) {\scriptsize Open-Vocabulary Detection};
    \node [anchor=north west,rotate=0]at (6.9,6.8) {\scriptsize Image Captioning};
    \node [anchor=north west,rotate=0]at (6.5,5.1) {\scriptsize Visual Question-Answering};
    \node [anchor=north west,rotate=0]at (1. ,6.9) {\scriptsize $x$};
    \node [anchor=north west,rotate=0]at (3.3,7) {\scriptsize $x'$};
\end{tikzpicture}
\caption{Another adversarial example created with PRM method using ViT-B/16 CLIP vision encoder as a surrogate. Clean ($x$) and adversarial ($x'$) inputs are shown in the top-left corner. Correct predictions are marked with green frames whereas adversarial predictions are marked with red. 
Curiously, as exemplified by this sample, OVD target models frequently make false positive predictions of \texttt{person}s, which is likely due to biases in the training data \cite{dataset-lin2014microsoftcoco} where \texttt{person} is the highest frequency class. }
\label{fig:qualitative-2010_002436}
\end{figure}
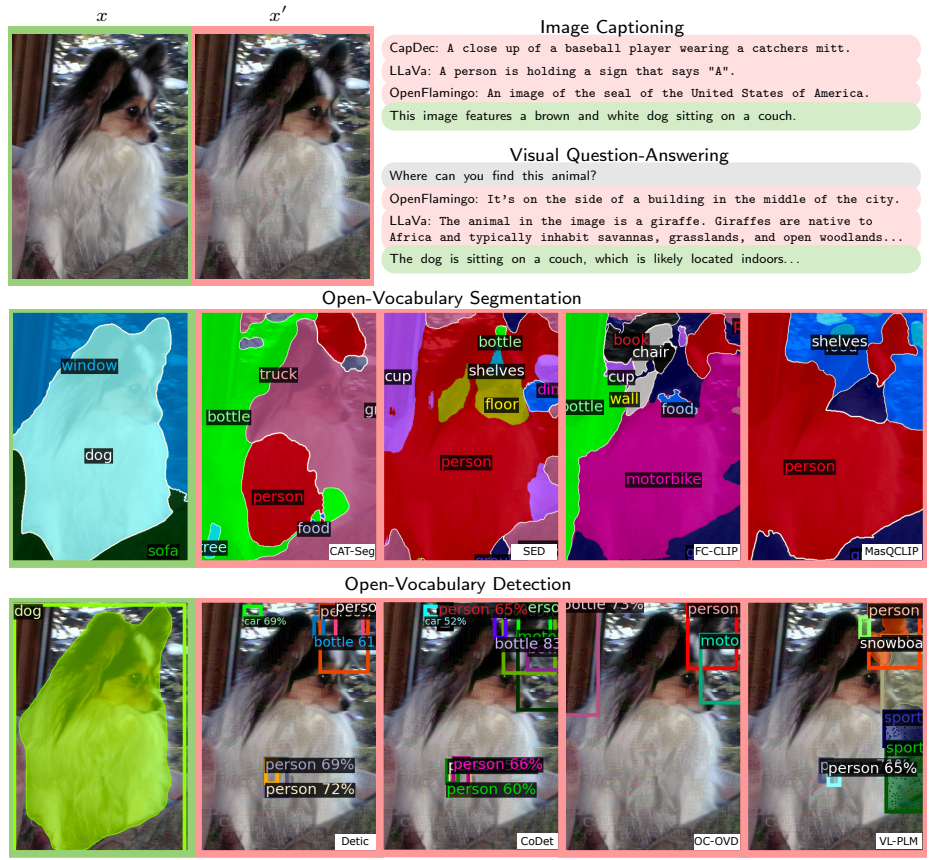

\begin{figure}[H]
    \centering
    \includegraphics[width=0.29\textwidth]{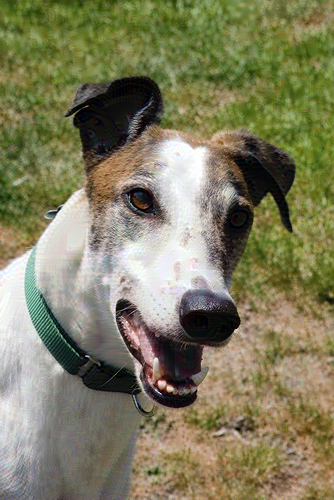}
    \includegraphics[width=0.29\textwidth]{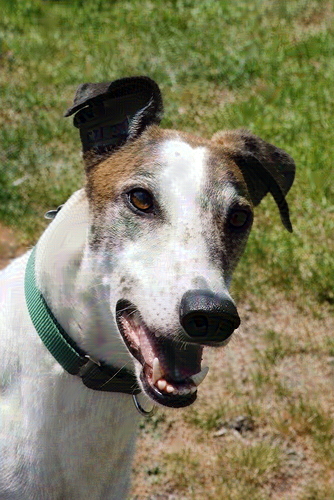}
    \includegraphics[width=0.29\textwidth]{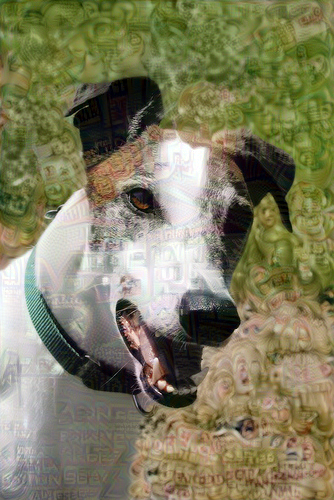}\\
    \includegraphics[width=0.29\textwidth]{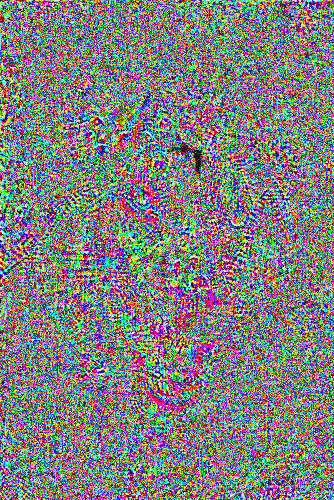}
    \includegraphics[width=0.29\textwidth]{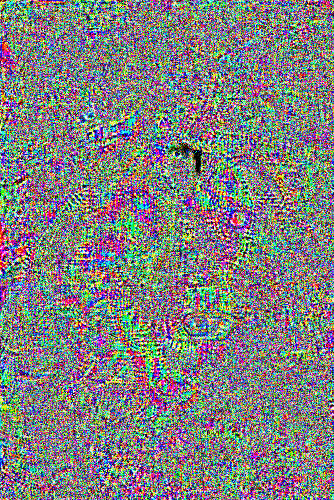}
    \includegraphics[width=0.29\textwidth]{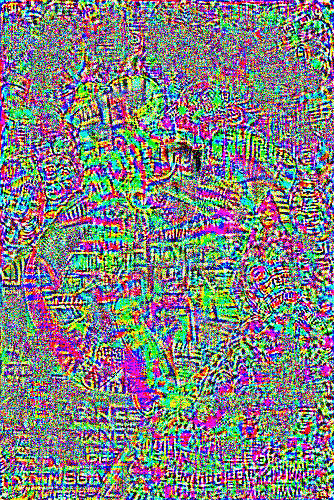}\\
    \includegraphics[width=0.29\textwidth]{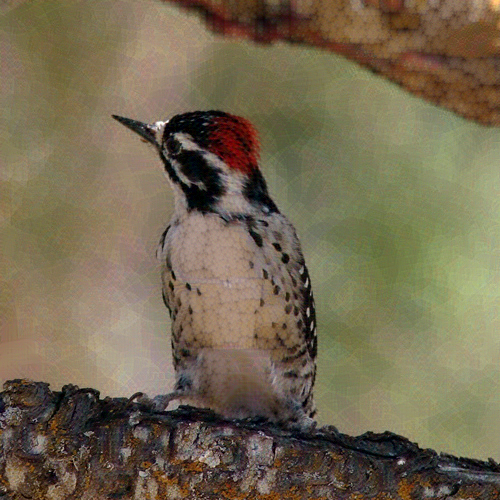}
    \includegraphics[width=0.29\textwidth]{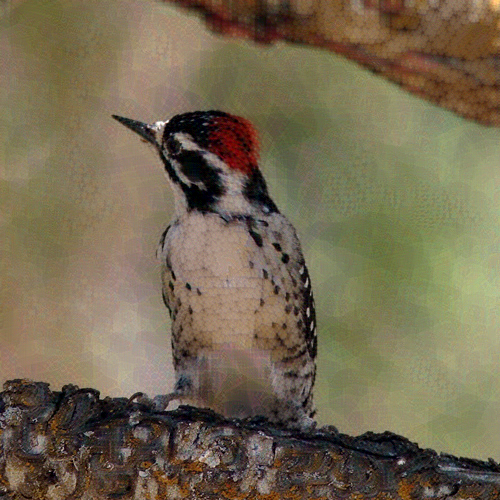}
    \includegraphics[width=0.29\textwidth]{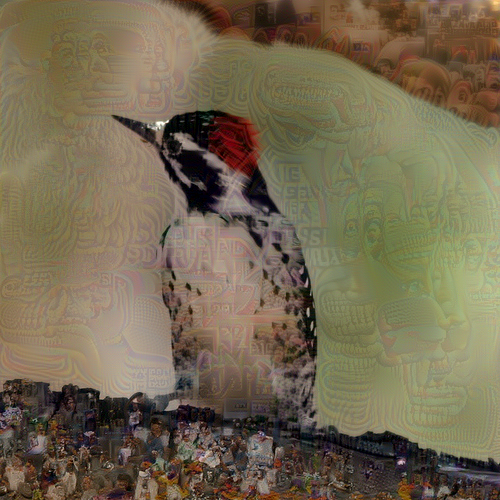}\\
    \includegraphics[width=0.29\textwidth]{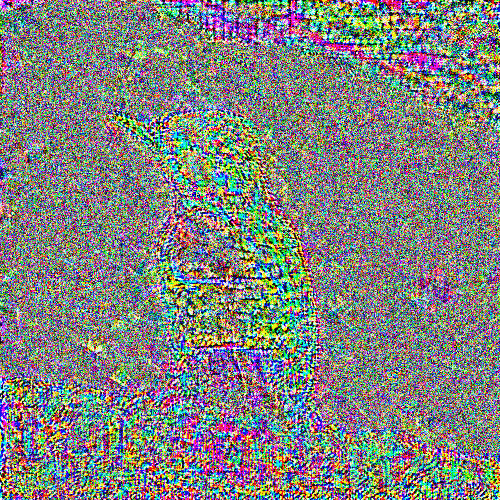}
    \includegraphics[width=0.29\textwidth]{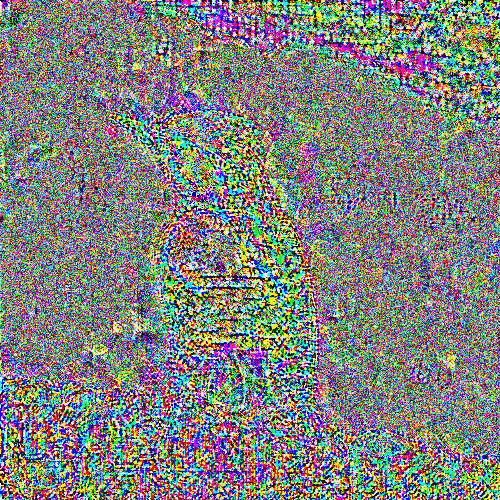}
    \includegraphics[width=0.29\textwidth]{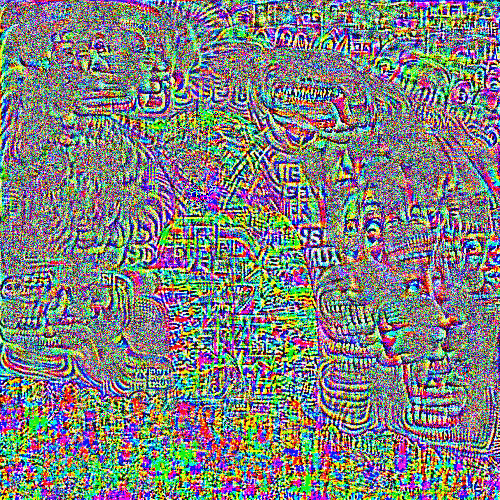}
    \caption{Comparison of adversarial perturbation patterns (magnified by 30$\times$) resulted from different attack strategies  (ViT-B/16 surrogate). Left to right: DR, NRD, PRM. Note that PRM can induce stronger patterned perturbations that cover the full canvas, including background and low spatial frequency image regions.}
    \label{fig:adv-patterns}
\end{figure}

\section{Additional Quantitative Results}\label{apx:quantitative}
\begin{table}[H]
\centering
\vspace{-20pt}
\fontsize{6}{8}\selectfont
\caption{Transfer attack efficacy on OVS target models. Model performance is measured in mean Intersection over Union (\textbf{mIoU}) on COCO-Stuff validation set.}\label{tab:ovs-coco}
\begin{subtable}{\textwidth}
\begin{tabular}{cc|C{1cm}|C{1.2cm}|C{1.2cm}|C{1.60cm}|C{1.60cm}|C{1.2cm}|C{1.2cm}}
\specialrule{.8pt}{1pt}{1pt}
 &  & \multicolumn{2}{c|}{SAN} & DeOP & ZSSeg-BL & ZegFormer & \multicolumn{2}{c}{CAT-Seg} \\
 & \multirow{-2}{*}{$\mathcal{L}_\mathrm{attack}$} & ViT-B & ViT-L & ViT-B & ViT-B & ViT-B & ViT-B & ViT-L \\
\multirow{-3}{*}{\backslashbox[8mm]{S}{T}} 
& Clean & 35.76 & 43.91 & 38.24 &  35.76 & 34.09 &  46.17 &  50.35   \\
 \hline \hline
SAN & $-\mathcal{L}_\mathrm{SAN}$ &\cellcolor[HTML]{EFEFEF}{\color{gray} 4.57} & 30.21 &\cellcolor[HTML]{EFEFEF}28.20 &\cellcolor[HTML]{EFEFEF}24.19 &\cellcolor[HTML]{EFEFEF}22.87 &\cellcolor[HTML]{EFEFEF}28.93 & 37.50 \\
DeOP & $-\mathcal{L}_\mathrm{DeOP}$ &\cellcolor[HTML]{EFEFEF}25.94 & 34.66 &\cellcolor[HTML]{EFEFEF}{\color{gray} 2.85} &\cellcolor[HTML]{EFEFEF}22.23 &\cellcolor[HTML]{EFEFEF}23.61 &\cellcolor[HTML]{EFEFEF}16.55 & 39.52\\
 \hline
 & DR &\cellcolor[HTML]{EFEFEF}38.46 & 42.24 &\cellcolor[HTML]{EFEFEF}35.29 &\cellcolor[HTML]{EFEFEF}32.21 &\cellcolor[HTML]{EFEFEF}31.22 &\cellcolor[HTML]{EFEFEF}42.07 & 48.13 \\
 & NRD &\cellcolor[HTML]{EFEFEF}26.44 & 37.80 &\cellcolor[HTML]{EFEFEF}24.33 &\cellcolor[HTML]{EFEFEF}26.45 &\cellcolor[HTML]{EFEFEF}26.39 &\cellcolor[HTML]{EFEFEF}27.25 & 44.23\\
\multirow{-3}{*}{\begin{tabular}[c]{@{}c@{}}CLIP\\ ViT-B\end{tabular}} & \textbf{PRM} &\cellcolor[HTML]{EFEFEF}\bluebf{3.89 }& \bluebf{12.85} &\cellcolor[HTML]{EFEFEF}\bluebf{3.48} &\cellcolor[HTML]{EFEFEF}\bluebf{7.06} &\cellcolor[HTML]{EFEFEF}\bluebf{7.47} &\cellcolor[HTML]{EFEFEF}\bluebf{4.23} & \bluebf{22.44} \\
 \hline \hline
 & DR & 37.64 & 40.82 & 32.65 & 27.08 & 25.61 & 38.91 & 46.18 \\
 & NRD & 36.67 & 39.65 & 30.68 & 24.63 & 23.81 & 39.06 & 45.17 \\
\multirow{-3}{*}{\begin{tabular}[c]{@{}c@{}}CLIP\\ {CNeXt-L}\end{tabular}} & \textbf{PRM} & \bluebf{26.88} & \bluebf{26.08} & \bluebf{12.10} &\bluebf{4.56} & \bluebf{5.09} & \bluebf{23.34} & \bluebf{32.09}\\
\specialrule{.8pt}{1pt}{1pt}
\end{tabular}
\\
\begin{tabular}{cc|c|c|C{.95cm}|c|c|c|C{.95cm}|C{.9cm}|C{.9cm}}
\specialrule{.8pt}{1pt}{1pt}
 &   & \multicolumn{2}{c|}{SED} & \multicolumn{3}{c|}{FC-CLIP} & SegCLIP & TCL & \multicolumn{2}{c}{SimSeg}\\
 \cline{3-11}
 & \multirow{-2}{*}{$\mathcal{L}_\mathrm{attack}$}& CNext-B & {CNeXt-L} & RN50 & RN50$\times$64 & {CNeXt-L}& {ViT-B} & {ViT-B} & {ViT-S} &{ViT-B} \\
 \cline{3-11}
\multirow{-3}{*}{\backslashbox[8mm]{S}{T}} 
& Clean & 46.21 & 49.57 &  54.85 &   60.53 &  63.22
 & 26.04 & 20.32 & 27.24 & 29.73\\
 \hline\hline
SAN & $-\mathcal{L}_\mathrm{SAN}$ & 32.50 & 39.34 & 42.43 & 52.14 & 55.28 &\cellcolor[HTML]{EFEFEF}13.12 &\cellcolor[HTML]{EFEFEF}14.09 & 19.53 &\cellcolor[HTML]{EFEFEF}21.20\\
DeOP & $-\mathcal{L}_\mathrm{DeOP}$  & 29.50 & 37.56 & 33.29 & 49.31 & 52.98 &\cellcolor[HTML]{EFEFEF}21.57 &\cellcolor[HTML]{EFEFEF}9.93 & 25.18 &\cellcolor[HTML]{EFEFEF}26.39\\
 \hline
 & DR & 41.92 & 46.46 & 49.20 & 57.14 & 59.93 &\cellcolor[HTML]{EFEFEF}26.20 &\cellcolor[HTML]{EFEFEF}20.32 & 27.13 &\cellcolor[HTML]{EFEFEF}29.23 \\
 & NRD & 34.88 & 42.53 & 39.43 & 52.46 & 55.52 &\cellcolor[HTML]{EFEFEF}20.14 &\cellcolor[HTML]{EFEFEF}10.05 & 24.28 &\cellcolor[HTML]{EFEFEF}26.23 \\
\multirow{-3}{*}{\begin{tabular}[c]{@{}c@{}}CLIP\\ ViT-B\end{tabular}} & \textbf{PRM} & \bluebf{11.57} & \bluebf{22.49} & \bluebf{13.25} & \bluebf{30.90} &  \bluebf{34.39} &\cellcolor[HTML]{EFEFEF}\bluebf{7.14} &\cellcolor[HTML]{EFEFEF}\bluebf{2.69} & \bluebf{18.31} &\cellcolor[HTML]{EFEFEF}\bluebf{19.98} \\
 \hline\hline
 & DR & 31.55 &\cellcolor[HTML]{EFEFEF}34.48 & 35.00 & 40.40 &\cellcolor[HTML]{EFEFEF}25.16  & 24.61 & 17.70 & 26.68 & 28.46\\
 & NRD & 27.41 &\cellcolor[HTML]{EFEFEF}29.51 & 32.06 & 37.78 &\cellcolor[HTML]{EFEFEF}19.60 & 23.97 & 17.12 & 26.21 & 28.48 \\
\multirow{-3}{*}{\begin{tabular}[c]{@{}c@{}}CLIP\\ {CNeXt-L}\end{tabular}} & \textbf{PRM} & \bluebf{4.35} &\cellcolor[HTML]{EFEFEF}\bluebf{5.65} & \bluebf{5.30} & \bluebf{7.07} &\cellcolor[HTML]{EFEFEF}\bluebf{0.62} & \bluebf{18.42} & \bluebf{11.73} & \bluebf{23.57} & \bluebf{25.72}\\
\specialrule{.8pt}{1pt}{1pt}
\end{tabular}
\end{subtable}
\end{table}

% \subsection{Rounding to the nearest Integer}

\section{Prompts used for IC and VQA evaluation}\label{apx:subsec:prompts}

Prompts used for caption generation:

\noindent OpenFlamingo:
{\fontsize{8}{7} \texttt{<example>Output:{\color{gray}\{EXAMPLE\}}...<image>Output:{\color{gray}\{CAPTION\}}<|endofchunk|>}}

\noindent LLaVA:
{\fontsize{8}{7} \texttt{USER: <image> Please describe the image using one sentence. ASSISTANT:}\vspace{1em}

\noindent Prompts used for VQA: 

\noindent OpenFlamingo: %\vspace{-20pt}
{\fontsize{8}{7} \texttt{<image> Question:{\color{gray}\{QUESTION\}} Short answer:{\color{gray}\{ANSWER\}} <|endofchunk|>}}

\noindent LLaVA: %\vspace{-20pt}
{\fontsize{8}{7} \texttt{USER:<image>{\color{gray}\{QUESTION\}} Answer the question using a single word or phrase.}}
{\fontsize{8}{7} \texttt{ASSISTANT:{\color{gray}\{ANSWER\}}}}

\section{Limitations and Broader Impacts}

\subsubsection{Limitations.} 
This work introduces a straightforward yet powerful attack strategy. We demonstrate how this approach can be used with an open-source model like CLIP to compromise downstream models built upon it. Our study provides substantial empirical evidence of such vulnerability inheritance issues in CLIP's downstream systems. We believe similar phenomena are likely present in downstream applications of other open-source foundational models, but leave further investigation for future research.

\noindent Our current attack focuses solely on CLIP's vision encoder, which may explain its limited effectiveness on VQA target systems compared to other tasks. Developing a stronger attack for tasks that have a greater emphasis on language (e.g. by involving the language modality in attack optimisation \cite{transfer-prompt-luo2023image} or explicitly leveraging contextual hints such as class co-occurrence information \cite{bg-adv-transfer-aich2022gama}) could be an interesting direction for future work.

\subsubsection{Potential Negative Societal Impacts.}
Our findings demonstrate the potential for using open-source foundation models like CLIP to compromise black-box downstream systems. While our research is presented with the goal of responsible disclosure, such findings may be exploited by malicious actors to compromise the integrity of real-world systems which could have negative societal impacts.

\end{document}